\documentclass[11pt]{article}
\usepackage{amsfonts}
\usepackage[final]{acl}
\usepackage{booktabs} 
\usepackage{times}
\usepackage{latexsym}
\usepackage{multirow}
\usepackage{amsmath}
\usepackage{multirow}
\usepackage{tcolorbox}

\usepackage{xcolor}
\usepackage{makecell}
\usepackage[labelformat=simple]{subcaption}

\usepackage[ruled,linesnumbered]{algorithm2e}

\usepackage[table]{xcolor} % 导入xcolor包并启用table选项
\usepackage[dvipsnames]{xcolor}

\usepackage[T1]{fontenc}
\usepackage[utf8]{inputenc}

\usepackage{microtype}

\usepackage{inconsolata}

\usepackage{graphicx}
\usepackage{url}

\newcommand{\thinkrow}[1]{\rowcolor{blue!10} \texttt{<think>} #1 \texttt{</think>} \\}
\newcommand{\searchrow}[1]{\rowcolor{green!10} \texttt{<search>} #1 \texttt{</search>} \\}
\newcommand{\resultrow}[1]{\rowcolor{gray!10} \texttt{<result>} #1 \texttt{</result>} \\}

\definecolor{deltaBg}{RGB}{220,230,255} % subtle blue-grey

\title{BAPO: Boundary-Aware Policy Optimization for Reliable Agentic Search}

\author{
Shiyu Liu$^{1,2,\dagger,\ddag}$, Yongjing Yin$^{2,\ddag}$, Jianhao Yan$^{4}$, Yunbo Tang$^{3}$, Qinggang Zhang$^{5}$\thanks{Corresponding author.}\\
{\bf Bei Li$^{2}$}, {\bf Xin Chen$^{2}$}, {\bf Jingang Wang$^{2}$}, {\bf Xunliang Cai$^{2}$}, {\bf Jinsong Su}$^{1,3,6}$\footnotemark[1] \\
$^{1}$Institute of Artificial Intelligence, Xiamen University
$^{2}$Meituan Inc.\\
$^{3}$School of Informatics, Xiamen University
$^{4}$Westlake University
$^{5}$Jilin University\\
$^{6}$Key Laboratory of Digital Protection and Intelligent Processing of Intangible Cultural \\ Heritage of Fujian and Taiwan (Xiamen University), Ministry of Culture and Tourism, China\\
\texttt{shiyu.liu0709@foxmail.com; qinggangzhang@jlu.edu.cn; jssu@xmu.edu.cn} 
}

\begin{document}
\maketitle

\insert\footins{\footnotesize$^\ddag$Equal contribution.}
\insert\footins{\footnotesize$^\dagger$Work done during internship at Meituan Inc.}

\begin{abstract}
RL-based agentic search enables LLMs to solve complex questions via dynamic planning and external search. While this approach significantly enhances accuracy with agent policies optimized via large-scale reinforcement learning, we identify a critical gap in reliability: these agents fail to recognize their reasoning boundaries and rarely admit ``\textit{I DON'T KNOW}'' (\textit{IDK}) even when evidence is insufficient or reasoning reaches its limit. The lack of reliability often leads to plausible but unreliable answers, introducing significant risks in many real-world scenarios. To this end, we propose \underline{\textbf{B}}oundary-\underline{\textbf{A}}ware \underline{\textbf{P}}olicy \underline{\textbf{O}}ptimization \textbf{(BAPO)}, a novel RL framework designed to cultivate reliable boundary awareness without compromising accuracy. BAPO introduces two key components: (i) a group-based boundary-aware reward that encourages an \textit{IDK} response only when the reasoning reaches its limit, and (ii) an adaptive reward modulator that strategically suspends this reward during early exploration, preventing the model from exploiting \textit{IDK} as a shortcut. Extensive experiments on four benchmarks demonstrate that BAPO substantially enhances the overall reliability of agentic search\footnote{Our code is available at \url{https://github.com/Liushiyu-0709/BAPO-Reliable-Search}.}.

\end{abstract}

\begin{figure}[t]
  \centering
  \includegraphics[width=1.0\linewidth]{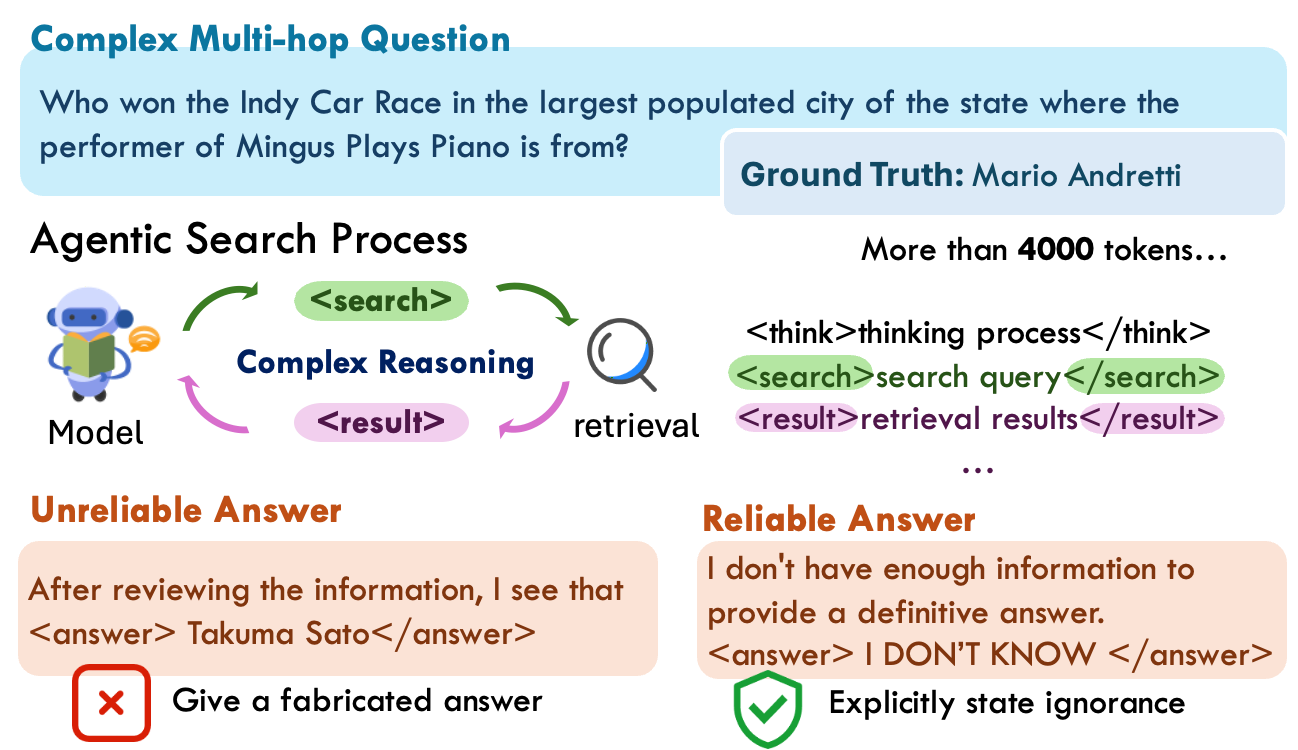} % 图片名/路径
  \caption{When the agentic search model produces wrong answers, its lengthy and complex reasoning makes it difficult for users to verify. To ensure reliability, the model should explicitly state when information is insufficient and that no answer is available.
}\vspace{-10pt}
  \label{fig:case}
\end{figure}
\section{Introduction}
% RL-based agentic search models, such as Search-R1~\cite{search-r1} and ReSearch~\cite{research}, are equipped to dynamically utilize search tools during their reasoning process for tackling complex, knowledge-intensive questions. 
Recent advances leverage reinforcement learning (RL) to optimize Large Language Models (LLMs) as autonomous agents that actively plan and execute multi-turn searches~\cite{search-r1,research,r1searcher}. 
While the RL-based agentic search substantially improves accuracy on complex, knowledge-intensive questions, it introduces a critical reliability issue: 
these RL-based models almost never admit \textit{``I DON'T KNOW''} (\textit{IDK}) even when evidence is insufficient or reasoning reaches its limit~\cite{ontheselfawareness_2025, sufficient_2025}. 
% reliability的重要性，
% A reliable model should not only solve problems accurately but also recognize when definitive answers are unavailable. 
As presented in Figure~\ref{fig:case}, when an agentic search model cannot reach a concrete final answer, it should explicitly admit \textit{IDK} rather than fabricating answers. 
This capability is especially valuable given the difficulty for users to verify the lengthy reasoning process, as honest acknowledgment of \textit{IDK} helps users seek alternative information sources and prevents misinformation.

Previous research attributes the ability to acknowledge uncertainty to a model’s awareness of its internal ``\texttt{boundary}''~\cite{knowledgeboundarylargelanguageli2025,ontheselfawareness_2025}. Specifically, EKBM~\cite{zheng2025enhancing} verifies information against its parametric \texttt{``knowledge boundary''} to answer factual questions accurately, while CapBound~\cite{ontheselfawareness_2025} introduces a similar concept to determine whether a problem is solvable within its \texttt{``capability boundary''} for mathematical reasoning. However, the reasoning boundary for an agentic search model is inherently more complex and dynamic, as such a boundary is continuously shaped by the interplay between the model's internal reasoning capacity for search tool interaction and the external retrieved information.

To investigate the impact of RL on an agentic search model's awareness of its reasoning boundaries, we conduct preliminary experiments measuring this awareness before and after RL training. A key finding is that while RL optimization improves accuracy on complex questions, it significantly degrades the model’s awareness of its boundary. This occurs because standard RL rewards incentivize exhaustive exploration for correctness while simultaneously discouraging any acknowledgment of uncertainty~\cite{search-r1,r1searcher}. 
As a result, these models often lack awareness of their boundaries and fabricate plausible answers.

Enhancing the boundary awareness of agentic search models is crucial but challenging. (i) It is difficult to quantitatively measure a model’s reasoning boundary, which makes it hard to construct a reliable reward signal for \textit{IDK} responses. 
Unlike static knowledge or capability boundaries, which are tied solely to the model's inherent competencies,
the boundary of an agentic search model is a dynamic, emergent property of the interaction between the agent's planning, the quality of its retrievals, and its iterative reasoning steps. Consequently, we lack a robust strategy to measure whether the agent has genuinely encountered its limit.
% This boundary is not known a priori and manifests only through the agent's own attempts to solve a problem. 
% Therefore, we lack a reliable metric to determine, at any point during training or inference, whether the agent has genuinely encountered its limit or is simply under-exploring.
(ii) It is challenging to integrate such a signal into the RL objective without creating unintended incentives.
A naive reward that encourages \textit{IDK} responses could be exploited as a low-effort shortcut, especially on difficult problems requiring deep exploration. If the reward function prioritizes \textit{IDK} over incorrect answers, the model may learn to default to \textit{IDK} prematurely, ultimately harming the reasoning accuracy and exploration depth.

To this end, we propose \underline{\textbf{B}}oundary-\underline{\textbf{A}}ware \underline{\textbf{P}}olicy \underline{\textbf{O}}ptimization \textbf{(BAPO)} for agentic search, designed to cultivate reliable self-awareness without compromising accuracy. 
Our research aims to address two fundamental research questions: (i) how to precisely model the learning signal for identifying the boundary of agentic search models, and (ii) how to integrate this signal into an RL framework to balance deep exploration and appropriate exploitation. 
% Extensive experiments show that BAPO achieves superior reliability, enabling the model to maintain awareness of its reasoning boundaries without compromising its capability for deep exploration. 
Results across four challenging benchmarks demonstrate that BAPO consistently outperforms mainstream training-based and prompt-based methods in overall reliability. Remarkably, with only $5000$ RL training samples, BAPO with Qwen2.5-7B-Instruct shows superior reliability compared to strong open-source agentic search models, without compromising average accuracy.

Our major contributions are listed as follows:
\begin{itemize}
\item We identify the key limitation of agentic search models and propose BAPO, a novel RL framework that dynamically rewards \textit{IDK} responses to cultivate the model's self-awareness of its reasoning boundaries.

\item BAPO introduces a group-based boundary-aware reward that encourages an \textit{IDK} response only when the question is out of the model's boundary, and an adaptive reward modulator to prevent reward hacking.

\item Extensive experiments on challenging benchmarks demonstrate that BAPO effectively enhances search reliability while preserving the model's ability to solve complex problems. 

\end{itemize}

\section{Background}
% 总结一下？

\subsection{Agentic Search}
To enhance response quality in knowledge-intensive scenarios, Retrieval-Augmented Generation (RAG) supplements LLMs' parametric knowledge with external search~\cite{rag_lewis2020, rag_gao2023, rag_fan2024}. Current prompting-based strategies often depend on predefined workflows for interleaving reasoning and search~\cite{selfrag, searcho1}, resulting in limited effectiveness. 
While learning-based methods (SFT/DPO) offer an alternative~\cite{ragit,corag, sft1_cognitivekernelproframeworkdeep, ragddr}, they remain constrained by complex data pipelines and suboptimal generalization.

Inspired by the success of reinforcement learning on mathematics and coding \cite{openaio1,deepseekai2025deepseekr1incentivizingreasoningcapability}, RL-based agentic search systems have emerged, such as Search-R1~\cite{search-r1}, ReSearch~\cite{research}, R1-Searcher~\cite{r1searcher}, and Tool-Star~\cite{toolstar_dong_2025}.
These methods employ format and outcome correctness rewards to enable autonomous query decomposition and flexible interaction with external search systems during reasoning, thereby achieving superior accuracy on complex queries.
Nevertheless, current methods prioritize search accuracy while ignoring holistic reliability, which depends on both accuracy and precision. 
In contrast, our proposed BAPO leverages boundary-aware reward mechanisms to optimize overall reliability, thereby maximizing the practicality of agentic search models.

\subsection{Formalization}
% To make it clear, we formalize the Agentic Search framework, which empowers an LLM to perform multi-hop reasoning and autonomous tool invocation to resolve complex queries.
% in this part

To make it clear, we formally define Agentic Search as a sequential decision-making process following the ReAct paradigm~\cite{react2023}. 
Given a query $x$, the agent aims to generate a trajectory $\tau$ that concludes with a final answer $y$. 
% In addition to the query, we typically include an system prompt to constrain the model's output format.
To structure this reasoning process, a system prompt is included to constrain the model's output format.
% define a set of potential tools $\mathcal{A}$ within the prompt and 
For instance, the model is required to use tags \texttt{<think>}, \texttt{<search>}, and \texttt{<answer>} to indicate thoughts, search tool calls, and the predicted answers, respectively, with search results returned in \texttt{<result>}.
The prompt template is detailed in Appendix~\ref{sec:tir-prompt}.
As formalized below, the trajectory $\tau$ is a sequence of interleaved reasoning steps, actions, and observations:
% $$\tau = (r_1, a_1, o_1, r_2, a_2, o_2, \dots, r_T, a_T, o_T, y),$$
% \begin{equation}
% \tau = (r_1, a_1, o_1, \dots, r_{T-1}, a_{T-1}, o_{T-1}, r_T,y),
% \tag{1}
% \end{equation}

\begin{equation}
\tau = \big( (r_t, a_t, o_t)_{t=1}^{T-1}, r_T, y \big),
\tag{1}
\end{equation}

\noindent where each $r_t$ denotes the reasoning state at step $t$ that plans subsequent actions and $r_T$ synthesizes accumulated information to conclude to the final answer $y$; 
% serving as the internal latent state where the model plans its next action.
$a_t$ is the search action; and $o_t$ is the returned search results.
% and $y$ is the final predicted answer.

% Following the ReAct paradigm \cite{react2023}, the policy $\pi_\theta$ maps the input $x$ and the preceding history $h_{<t} = (x, r_1, a_1, o_1, \dots, o_{t-1})$ to a distribution over the next reasoning trace and action:
% \begin{equation}
%     P(r_t, a_t | h_{<t}) = \pi_\theta(r_t, a_t | h_{<t}).
% \end{equation}
% The transition to the next state is governed by the intermediate search results. The process continues until the model emits a termination token within $a_T$ or reaches a maximum depth $T$, at which point the final answer $y$ is generated based on the full context: $P(y | h_{\leq T})$.

As RL has demonstrated significant efficacy in enhancing LLM reasoning capabilities, 
% we adopt Group Relative Policy Optimization (GRPO)~\cite{deepseekmath_shao_2024} as our learning algorithm, which is extensively employed in training agentic search frameworks~\cite{search-r1, research,toolstar_dong_2025}.
Group Relative Policy Optimization (GRPO)~\cite{deepseekmath_shao_2024} has been extensively employed in training agentic search models~\cite{search-r1, research,toolstar_dong_2025}.
% The model is optimized by RL algorithms (e.g, Group Relative Policy Optimization (GRPO) \cite{deepseekmath_shao_2024}), which allows the model to iteratively refine its reasoning traces and search strategies.
For a query $x$, a group of $G$ trajectories $\{\tau_1, \tau_2, \dots, \tau_G\}$ is sampled from the policy, and the objective function is defined as:
\begin{align*}
&\mathcal{J}(\theta)= \mathbb{E}_{x \sim \mathcal{D}, \{\tau_i\}_{i=1}^G \sim \pi_{\theta_{\text{old}}}(\cdot|x)} \tag{2}\frac{1}{G} \sum_{i=1}^{G} \\
&\left[ \min \left(w_i(\theta) A_i, \text{clip}\left(w_i(\theta), 1-\epsilon, 1+\epsilon\right) A_i \right)\right],
% & w_i(\theta)=\frac{\pi_{\theta}(\tau_i \mid x)}{\pi_{\theta_{\text{old}}}(\tau_i \mid x)},
% A_i = \frac{r_i - \text{mean}(\{r_j\}_{j=1}^G)}{\text{std}(\{r_j\}_{j=1}^G)}.
\end{align*}
where $w_i(\theta)=\frac{\pi_{\theta}(\tau_i \mid x)}{\pi_{\theta_{\text{old}}}(\tau_i \mid x)}$ and the KL term against a reference model is omitted.
% Moreover, a KL divergence penalty $\mathbb{D}_{KL}(\pi_\theta \| \pi_{\theta_\text{ref}})$ with $\beta$ as a coefficient can be added to the objective to prevent the policy from deviating too much from the original reference policy.
The advantage $A_i$ is computed by normalizing the rewards within the group, $A_i=\frac{\mathcal{R}(\tau_i) - \text{mean}(\{\mathcal{R}(\tau_j)\}_{j=1}^G)}{\text{std}(\{\mathcal{R}(\tau_j)\}_{j=1}^G)}$.
% \begin{equation}
%     A_i = \frac{\mathcal{R}(\tau_i) - \text{mean}(\{\mathcal{R}(\tau_j)\}_{j=1}^G)}{\text{std}(\{\mathcal{R}(\tau_j)\}_{j=1}^G)}.
% \tag{3}
% \end{equation}

Rule-based rewards have demonstrated robust empirical performance 
and are widely adopted in current research \cite{r1searcher, research}. 
Typical reward design focuses on correctness and consists of two parts: (i) a format correctness reward that checks whether the output conforms to the prescribed structure across reasoning steps, tool calls, and answers; and (ii) an outcome correctness reward that measures the objective correctness of the final answer.
% These reward designs typically comprise two components: 
% (i) a format correctness reward that evaluates whether the model's output conforms to the prescribed structure, encompassing reasoning thoughts, tool calls, and answers and (ii) an outcome correctness reward that measures the objective correctness of the final answer.
% Specifically, the format correctness reward evaluates whether the model's output conforms to the prescribed structure, encompassing reasoning thoughts, tool calls, and answers, whereas the outcome correctness reward measures the objective correctness of the final answer:
\begin{equation}\label{eq:factual-score}
\mathcal{R}^\textit{Correct}=
\begin{cases}
\displaystyle F1(\hat{y}, y),& \textit{correct format}\\[6pt]
-1, & \textit{wrong format} ,\tag{3}
\end{cases}
\end{equation}
where F1 means character-level F1 score between two strings, $y$ is the prediction answer extracted from trajectory and $\hat{y}$ is the ground truth answer.

\section{Preliminary Study}
In this study, we first investigate the impact of RL training with correctness-based reward on the boundary awareness of agentic search models. Subsequently, we incorporate an additional reward term designed to encourage \textit{IDK} responses and evaluate its effect. Our findings reveal an inherent trade-off between maximizing accuracy and maintaining effective boundary awareness.
% Experimental settings are elaborated in Section~\ref{sec:pre-exp}.
% performance of the resulting model reveals the challenges associated with the trade-off between accuracy and boundary-awareness during RL.

% we examine the 带来的影响 of incorporating a new reward of encouraging \textit{IDK}, and reveal the challenges associated with the trade-off between accuracy and boundary-awareness during RL.

% \subsection{RL with correctness reward impairs model's awareness of boundary}
\subsection{Setup}
\label{sec:settings}
\paragraph{Training.} 
For training, we use a high-quality RL training dataset from \citet{toolstar_dong_2025}, comprising 5000 multi-hop QA samples from HotpotQA and 2WikiMultihopQA. 
Following the cold-start-free paradigm, we directlt apply GRPO to Qwen2.5-3B-Instruct~\cite{qwen2025qwen25technicalreport}, bypassing preliminary in-domain SFT.
% The reward function jointly assesses outcome correctness and format adherence. 
% The training dataset is derived from \citet{toolstar_dong_2025}, consisting of 5,000 examples .

\paragraph{Evaluation.}
For evaluation, we use four multi-hop QA benchmarks: HotpotQA~\cite{hotpotqa}, MuSiQue~\cite{musique}, 2WikiMultiHopQA~\cite{2wiki}, and Bamboogle~\cite{bamboogle}. Specifically, HotpotQA, 2WikiMultiHopQA and MuSiQue are constructed from Wikipedia or Wikidata~\cite{wikidata} using diverse multi-hop mining strategies, while Bamboogle is a manually curated dataset of 2-hop challenging questions.
Following the setup in \citet{toolstar_dong_2025}, we construct a validation set of 80 multi-hop QA pairs, comprising 20 examples from the test set of each benchmark.
% the reported results in this section use 80 multi-hop QA pairs, including 20 examples each from each of the test sets, 
In contrast, the results reported in Section~\ref{sec:experiments} are evaluated on the complete test sets.

\paragraph{Metrics.}
To comprehensively evaluate model reliability, we follow \citet{rejectionIR} to use three following metrics: accuracy (\text{acc}), precision (\text{prec}) and \textit{IDK} rate ($\rho_{\textit{IDK}}$). These metrics are defined as follows:
\begin{equation}
\text{acc} = \frac{N_c}{N},\text{prec}=\frac{N_c}{N-N_r}, \rho_{\textit{IDK}}=\frac{N_r}{N},
\tag{4}
\end{equation}
where $N=N_c+N_w+N_r$ is the total number of answers, with $N_c$, $N_w$, $N_r$ denoting the number of correct answers, wrong answers and \textit{IDK} answers, respectively. Following previous studies~\cite{research, toolstar_dong_2025}, we extract answers from the model output enclosed in $\textbackslash \text{box}\{\}$, and then use GPT-4 as a judge to assess the correctness (Appendix~\ref{sec:llm-judge-template}).

% To evaluate the correctness of answers, we use LLM-as-a-judge Equal~\cite{llmjudge1} for automatic evaluation as it has a more flexible evaluation criterion to avoid misclassifying semantically equivalent but lexically mismatched answers as incorrect. 

% \paragraph{RL Training Setting.} 

\begin{figure}[t]
  \centering
  \includegraphics[width=.95\linewidth]{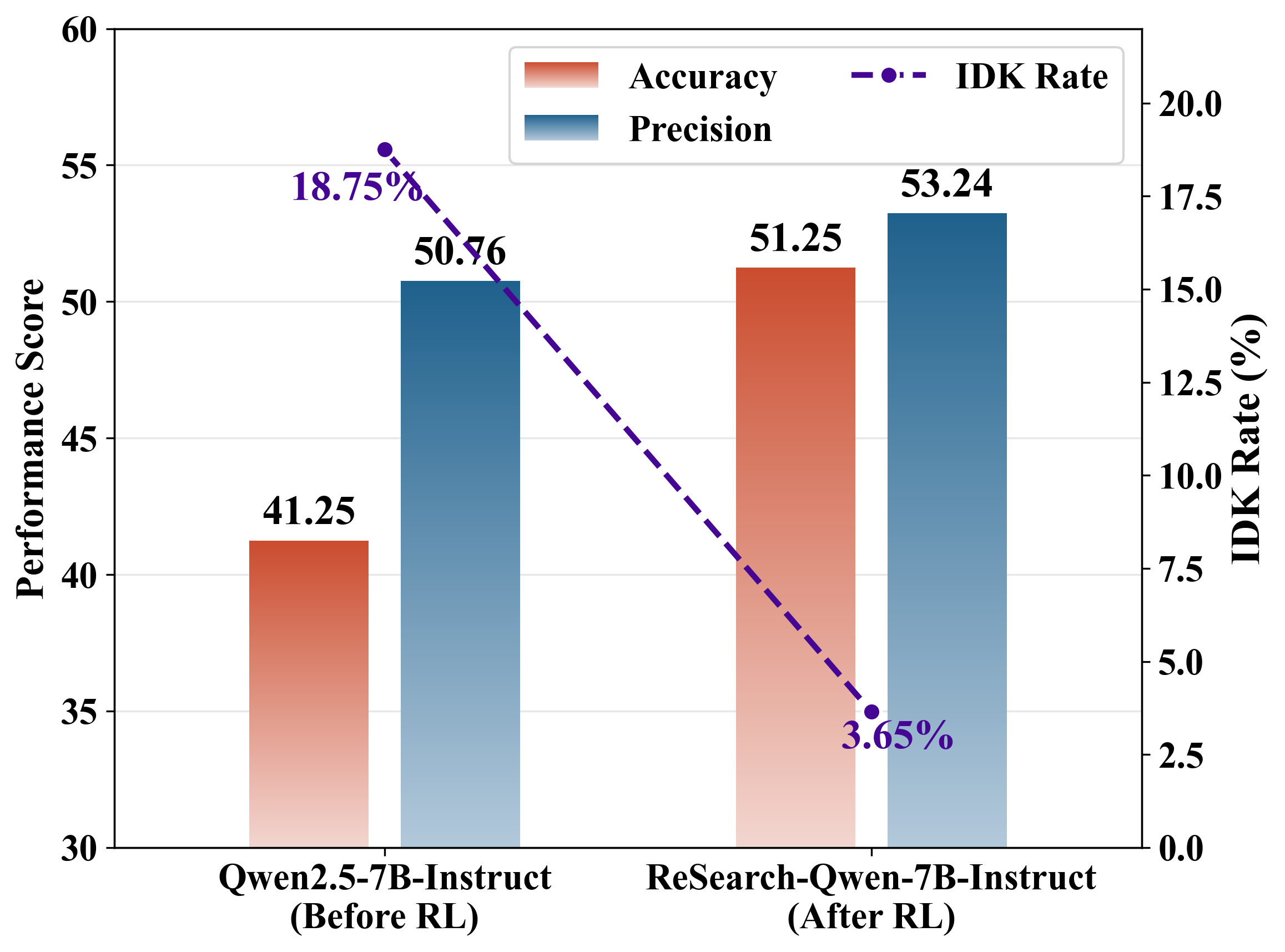} % 图片名/路径
  \caption{Evaluation results of accuracy, precision and \textit{IDK} rate ($\rho_{\textit{IDK}}$) of models before and after RL. The sharp drop in $\rho_{\textit{IDK}}$ coupled with the narrowing gap between accuracy and precision, indicates a diminished boundary awareness after RL.}\vspace{-5pt}
  \label{fig:boundary}
\end{figure}

%左边的validation score改成validation accuracy, original改成vanilla.
\begin{figure}[t]
  \centering
  \includegraphics[width=1.0\linewidth]{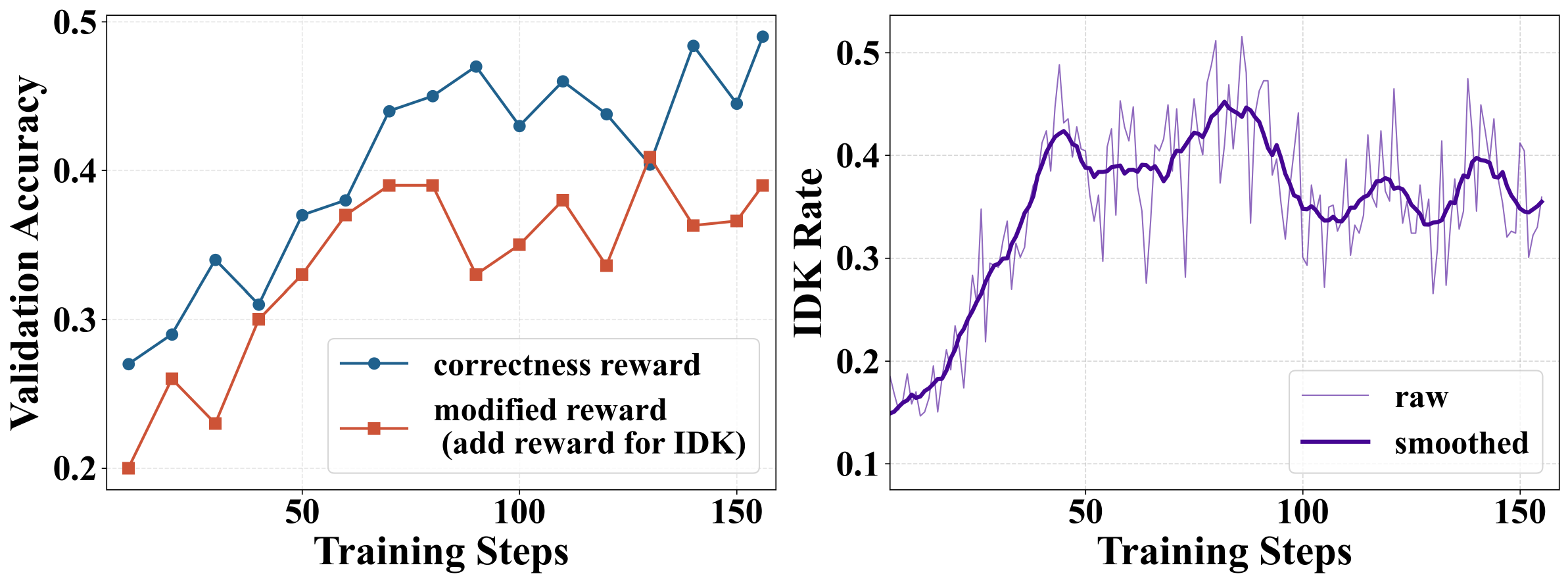} % 图片名/路径
  \caption{\textbf{Left:} Validation accuracy under different reward settings during RL training. \textbf{Right:} \textit{IDK} rate ($\rho_{\textit{IDK}}$) under the modified reward during RL training.
  % modified reward下的\textit{IDK} rate ($\rho_{\textit{IDK}}$) 的动态变化.
  % The validation f1 score during training process under two reward settings with Qwen2.5-3B-Instruct.
  }
  \vspace{-13pt}
  \label{fig:rba_only}
\end{figure}

% Accuracy measures models' overall problem-solving ability, while precision measures the correctness of provided (non-rejected) answers.
% 1.这里我们采用GRPO算法代表RL算法，进行实验。训练样本由5000条包含Hotpotqa以及2wiki的training set部分的数据数据组成，followingxxx,xxx，GRPO对应的reward是基于outcome的结果正确性和格式正确性的。
% 2.我们的验证集是由80条来自HotpotQA, Musique, 以及2Wiki的Bamboogle的数据组成，每个数据集分别20条，多样的数据来源帮助我们综合衡量模型的performance。
\begin{figure*}[t!]
  \centering
  \includegraphics[width=1.0\textwidth]{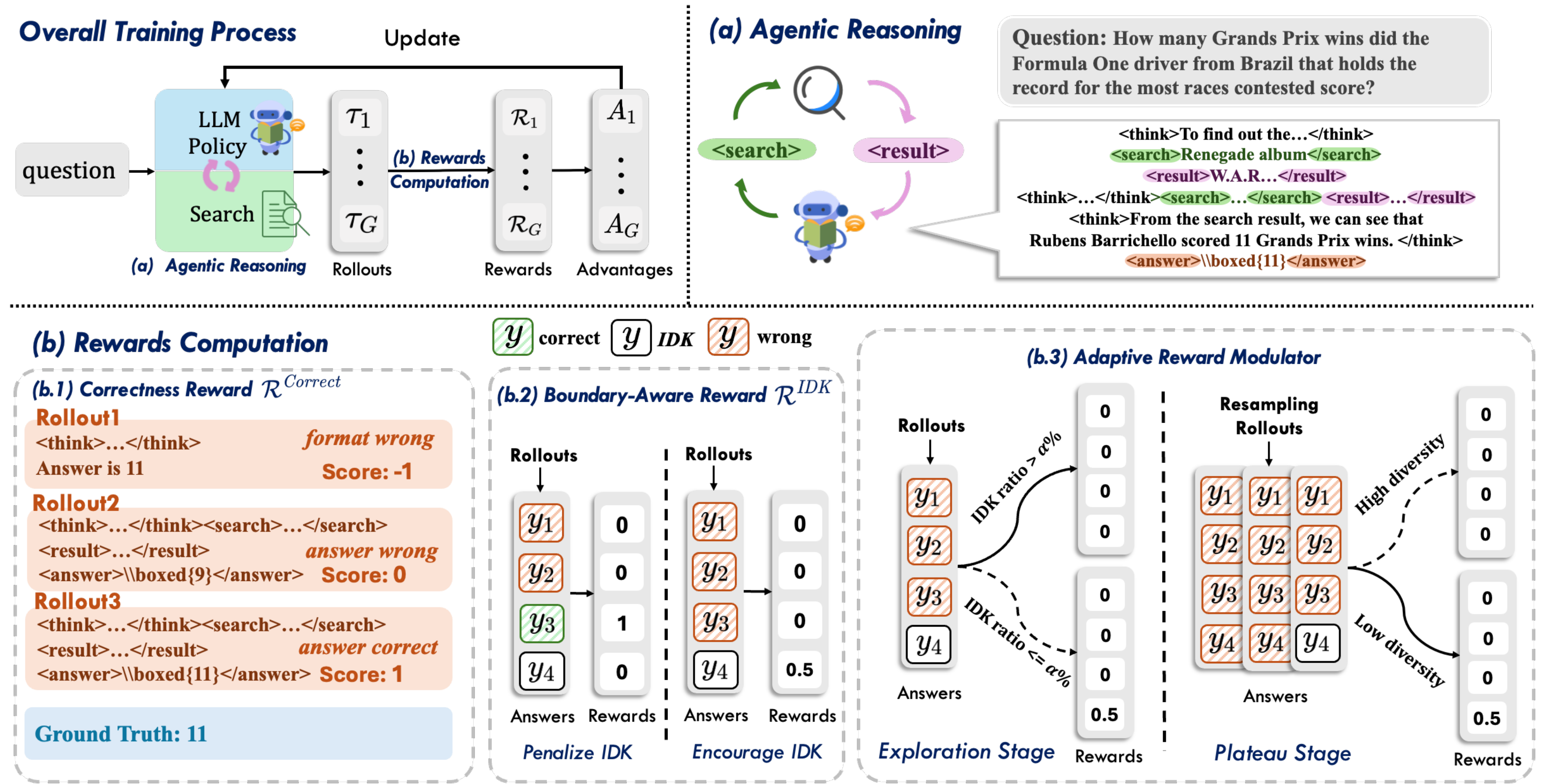}
  \caption{The overall framework of BAPO. Its training process begins with \textbf{(a)} \textbf{agentic reasoning},
  % over training instances, 
  where the model generates a group of rollouts for each question by interleaving thought processes with search environment interactions. These rollouts are then passed to the \textbf{(b)} \textbf{reward computation} module, which is comprised of \textbf{(b.1)} \textbf{correctness reward $\mathcal{R}^{\textit{Correct}}$} derived from the correctness of format and outcome, \textbf{(b.2)} \textbf{boundary-aware reward $\mathcal{R}^{\textit{IDK}}$} designed to incentivize \textit{IDK} responses when no correct rollout exists within the group, and \textbf{(b.3) \textbf{adaptive reward modulator}} adaptively disabling $\mathcal{R}^{\textit{IDK}}$ based on \textit{IDK} ratio at the exploration stage and diversity of rollouts at the plateau stage. 
  % Each training step concludes with a policy update based on advantages.
  % maintain the model's exploration during RL, a reward modulator
}
\vspace{-10pt}
  % The overall training process of BAPO. 
  % Given a question, the model first generates a group of rollouts with (a) agentic reasoning. through model's interaction with search environment during reasoning, the model finally conclude to a final answer.
  % Then, each rollout is assigned a reward based on its outcome through rewards computation process. BAPO's rewards composed of (b.1) basic rewards: decied by 单条rollout本身的 format 合法性 and answer correctness and (b.2) Reaosoning-boundary Aware Reward, which assigns a reward of 0.5 to encourage \textit{IDK} response when no correct rollouts exists in the group. During exploration stage and plateau stage, reward for \textit{IDK} rollouts are adaptively deactivated to maintains model's exploration.
  % 最后，
  % then group-relative advantages are computed for updating model. For simplicity, we omit the KL regularization. 
  
  % \textbf{(a)} Agentic reasoning with the external search. \textbf{(b)} Training Process: given a question, the model generates a group of rollouts. Each rollout is assigned a reward and then group-relative advantages are computed for updating model. For simplicity, we omit the KL regularization. 
  % \textbf{(c)} Basic reward is decided by the answer correctness and format. \textbf{(d)} Reasoning-Boundary Aware Reward: when correct answer exists in rollouts, zero reward is assigned to \textit{IDK}. During exploration stage and plateau stage, reward for \textit{IDK} rollouts are adaptively deactivated to maintains model's exploration.
    \label{fig:method}
\end{figure*}

% \subsection{RL with Correctness Reward Impairs Model's Reliability of Acknowledging \textit{IDK}}
\subsection{RL with Correctness Reward Impairs Boundary Awareness}
\label{sec:impairs}
To evaluate the impact of RL on boundary awareness, we compare Qwen2.5-7B-Instruct with its variant, ReSearch-Qwen-7B-Instruct~\cite{research}, which is optimized via GRPO using a correctness-based reward. To elicit explicit \textit{IDK} admissions, 
we follow \citet{rejectionIR} to append the following reliable prompt to the original one:
% we adopt a reliable prompting strategy from \citet{rejectionIR}, appending the following instruction to the standard agentic search prompt:
\begin{tcolorbox}[colback=gray!10, colframe=black!100, title=Reliable Prompt]
\texttt{If you can't solve this question by the reasoning process, you should output \textbackslash boxed\{I DON'T KNOW\}.}
\end{tcolorbox}

% 除了衡量models在reliable prompt下的表现之外，我们还引入the original prompt 
% \noindent We also incorporate the standard prompt for comparison. Note that under the standard prompt, precision equals accuracy since the model is not instructed to abstain from providing answers.

As illustrated in Figure~\ref{fig:boundary}, before RL training, Qwen2.5-7B-Instruct shows a notable precision advantage over its accuracy (50.76 vs. 41.25), accompanied by a substantial \textit{IDK} rate of 18.75\%, effectively filtering uncertain cases.
However, after RL training (ReSearch-Qwen-7B-Instruct), while accuracy improves to 51.25, \textit{IDK} rate drops sharply to 3.65\%, and precision increases only marginally to 53.24. 
This indicates a weakened boundary awareness that suppresses \textit{IDK} admission.
% This significant reduction in \textit{IDK} admissions—despite limited precision gains—suggests a diminished boundary awareness in the model's decision-making. The RL process appears to suppress the model's ability to recognize and abstain from uncertain cases, potentially leading to overconfidence in its predictions.

% As shown in Figure~\ref{fig:boundary}, prior to RL, Qwen2.5-7B-Instruct achieves higher precision(这里更高的precision是相对于它的accuracy而言的) and a substantial \textit{IDK} rate of 18.75\%, effectively rejecting uncertain cases. However, after RL (ReSearch-Qwen-7B-Instruct), although accuracy increases, the \textit{IDK} rate drops sharply to 3.65\% and the improvement in precision is marginal. This trend suggests diminished boundary awareness that suppresses \textit{IDK} admission.

% As shown in Figure~\ref{fig:boundary}, prior to RL (Qwen2.5-7B-Instruct), employing the reliable prompt yields a higher precision and a substantial \textit{IDK} rate of 18.75\%, indicating effective rejection of uncertain cases. However, after RL (ReSearch-Qwen-7B-Instruct), although accuracy further increases, the \textit{IDK} rate declines sharply to 3.65\% and the improvement in precision is marginal, implying a diminished boundary awareness that suppresses \textit{IDK} admission.

% revealing that the RL-trained model is less likely to admit \textit{IDK}, suggesting a diminished boundary awareness.

\subsection{Encouraging \textit{IDK} Responses Hinders Accuracy Advancement}
\label{sec:hinder}
Since correctness-based rewards alone weaken the model's awareness of when to acknowledge IDK, a natural approach is to incorporate appropriate positive rewards for \textit{IDK} responses during RL training.
Following the training setting in Section~\ref{sec:settings}, we instruct the model with reliable prompt and modify the correctness reward function by assigning an additional reward of 0.5 to \textit{IDK} responses within rollout groups that lacks correct answers. This design prioritizes \textit{IDK} responses over incorrect answers when a correct answer is unavailable.
% with the expectation of encouraging the model to develop boundary awareness.

% We assume the reason that models after RL struggles to exhibit reliability lies in the misalignment between the reward function in RL and user's expectation for model's reliability.

However, as shown in Figure~\ref{fig:rba_only}, directly incentivizing \textit{IDK} responses leads to reward hacking: the model learns to maximize rewards by defaulting to \textit{IDK} rather than attempting to solve challenging problems. Consequently, this modified reward impedes accuracy gains compared to the vanilla reward. This finding underscores a fundamental challenge in training reliable models: balancing the model's exploration for correctness rewards against its exploitation for \textit{IDK} rewards.

%我们的贡献可以总结为以下几点：
%1.我们揭示了现有RL算法正确性的奖励
%2.提出了Reward
%3

%Changllge
% To improve model reliability, reliable prompting has been 证实是一个简单有效的方法，however, when we apply it int
% Therefore, 我们用the reasoning boundary to define the 可得性 of a final answer that can be supported by the information obtained within the interleved reasoning-retrival trajectory.
% In contrast, the boundary of agentic search models is not only depends on model's predefined reasoning ability, but also relevant to the retrived information during reasoning. therefore, 我们定义了模型的reasoning boundaries as model已有的reasoning trajetory中的得到的信息能得出的支撑的结论范围。

% instead, the model assesses its ability to answer a question only after executing its reasoning trajectory and engaging with retrieved content. This means their effective capability boundary is determined dynamically during the reasoning process itself.

% When adopting reliable prompting to Qwen2.5-7B-Instruct, the model achieves higher precision while at the minor cost of lower accuracy when compared to the orginal prompt. After the model's training on GRPO, the reliable prompt can't bring enough boost to the reliability.

\section{The Framework of BAPO}
% 为了探究如何改善Agentic Search系统的可靠性，我们从基础的reliable prompt策略激发模型对自身推理边界的意识，并进一步揭示了基于结果正确性的奖励会损害模型的这种自我意识到推理边界的能力。为此，我们在GRPO的基础上，提出了基于group的推理边界意识奖励。无需任何额外的数据构造，在模型解决问题能力的变化中去动态决定推理边界。
% We begin with a basic reliable prompting strategy and conclude that while instruct model's awareness of its own reasoning boundaries can be activated, correctness-based reward of RL impairs this self-awareness capability.
% 基于我们如上的发现，我们决定通过

With observations and insights from the preliminary study, we propose Boundary-Aware Policy Optimization (BAPO), an RL algorithm built upon GRPO for training reliable agentic search models. 
As shown in Figure~\ref{fig:method}, BAPO uses boundary-aware reward to encourage
\textit{IDK} responses according to whether a problem is out of model's boundary. 
More importantly, it uses an adaptive reward modulator to balance exploration and exploitation, mitigating the reward hacking issue identified in Section~\ref{sec:hinder}.

\subsection{Boundary-Aware Reward}
We regard that a question exceeds the model's boundary if the model fails to get any correct answer among multiple rollouts.
% , we adjust the reward signal for \textit{IDK} responses.
% We have demonstrated that as the size of rollouts increases, the boundaries can be approximately determined by whether the model fails to get any correct answer among multiple rollouts.
In this situation, we adjust the reward assignment for \textit{IDK} responses.
Formally, for a set of rollout outputs $\{\tau_i\}_{i=1}^G$ and their corresponding correctness rewards $\{\mathcal{R}^\textit{Correct}_i\}_{i=1}^G$, the boundary-aware reward is calculated as:
% \begin{equation}
% \label{eq:factual-score}
% r_{\mathrm{BA}}=
% \begin{cases}
% \displaystyle 0.5, & all r_{\mathrm{correct}} <= 0,\\[6pt]
% 0, & eixst r_{\mathrm{correct}} > 0.
% \end{cases}
% \end{equation}
% 这一奖励只apply到I Don't Know的output, 当一组中的正确性奖励符合$\forall r_{\mathrm{correct}} > 0$时，$r_{\mathrm{rba}} = 0.5$,否则$r_{\mathrm{rba}} = 0$.
% \begin{equation} \small
% \label{eq:factual-score}
% \mathcal{R}^\textit{IDK}=\displaystyle 0.5 \cdot \mathbb{I}(\hat{y}_i=\textit{IDK}), \\ \text{if} \;\;\forall \{\mathcal{R}^\textit{Correct}(\tau_i)\}_{i=1}^G \leq 0. \tag{6}
% \end{equation}
\begin{equation} 
\label{eq:factual-score}
\begin{split}
&\mathcal{R}^\textit{IDK} = \displaystyle 0.5 \cdot \mathbb{I}({y}_i=\textit{IDK}),\\ 
&\text{if} \;\;\forall \{\mathcal{R}^\textit{Correct}(\tau_i)\}_{i=1}^G \leq 0.
\end{split} \tag{5}
\end{equation}
% This dynamic reward allocation mechanism is based on a simple assumption: If no correct output exists in a group of rollouts, the problem is deemed to exceed the model's reasoning boundary, and the model should be encouraged to admit \textit{IDK}. Otherwise, the model should be encouraged to generate reasoning paths toward correct answers.
% This group-based design ensures the model to establish a reasonable awareness of its reasoning boundary. 
\noindent The final reward $\mathcal{R}$ is the sum of $\mathcal{R}^\textit{Correct}$ and $\mathcal{R}^\textit{IDK}$:
\begin{equation}
\label{eq:final-score}
\mathcal{R}=\mathcal{R}^\textit{IDK} + \mathcal{R}^\textit{Correct}.
\tag{6}
\end{equation}

% \subsection{Exploration-Boundary Tradeoff}
% \subsection{Adaptive Reward Mechanism}
\subsection{Adaptive Reward Modulator}

% However, directly applying the Reasoning-Boundary Aware Reward $r_{\mathrm{rba}}$ to RL training creates a conflict between the model's problem-solving capability and its reliability. Specifically, the model learns to maximize rewards by giving more \textit{IDK} responses rather than attempting to solve harder problems, which ultimately impairs its exploration of challenging problems. 
% However, directly applying $r_{\mathrm{rba}}$ to RL training creates a shortcut to get more reward by giving more \textit{IDK} responses rather than attempting to solve harder problems, which ultimately impairs model's exploration of challenging problems. 
% To balance the enhancement of model's problem solving and their boudnary-aware ability.

Our preliminary analysis suggests that purely incentivizing \textit{IDK} responses leads to a degradation in exploration. 
To balance exploration and boundary awareness during RL training, we introduce an adaptive reward modulator for  $\mathcal{R}^{\textit{IDK}}$, which functions at two distinct levels:
\paragraph{Stage Level.}
In the early stages of training, the model needs to conduct extensive exploration to learn how to solve tasks. During this \textbf{exploration stage}, we avoid introducing additional rewards to interfere with the model's learning. $\mathcal{R}^\textit{IDK}$ is deactivated by default and only activated when the proportion of \textit{IDK} responses $\mathcal{R}^\textit{IDK}$ falls below $\alpha$.
As training progresses, the task accuracy gradually stabilizes.
At this \textbf{plateau stage}, we apply the reward $\mathcal{R}^\textit{IDK}$ fully to make the model start paying more attention to its boundary awareness.
Some difficult queries may not be easily answered correctly or admitted as unknown under the default group size, and thus we dynamically resample groups with no correct rollouts up to $k$ times, until the model outputs \textit{IDK} or provides a correct answer.
The transition between the exploration and plateau stages is determined by the stagnation of the validation score, e.g., when the validation score plateaus for 5 consecutive steps.
\paragraph{Sample Level.}
The consistency of a model's generated responses can serve as a proxy for its confidence.
    To further preserve the model's exploration during the plateau stage, 
    we adaptively modulate $\mathcal{R}^\textit{IDK}$ based on the rollout diversity of each query.
    Specifically, for queries exhibiting high rollout diversity, 
    indicating that the model is actively exploring the solution space, we deactivate $\mathcal{R}^\textit{IDK}$ to prevent untimely convergence.
    In contrast, for samples with low rollout diversity, which signifies that the model has converged on a specific output, $\mathcal{R}^\textit{IDK}$ is applied to refine its boundary awareness. Formally, a sample is considered to have high rollout diversity if its corresponding rollout answers $|\{{{y}_{1...G}}\}| \ge \frac{G}{2}$; otherwise, it is categorized as having low diversity.

% However, when directly apply the boundary-aware reward into RL training. A conflict between model's ability for answering question and model's reliability happens. The model learn's to 通过回答更多的IDK来获得奖励 rather than 尝试去回答更难的问题以期望得到basic奖励. this situation impairs model's exploration on harder problems. Therefore, we propose regularization strategy for Boundary-Aware Rward for avoid imparing model's exploration while achieve models' boundary-awareness. The strategy is composed of two aspects, the training stage and rollout structure.
% For training stage, as model tends to explorate at the early stage, we control the IDK reward only activated when the IDK ratio is lower than 5\%. and in later stage that model's performance platue, we takes the Boundary-Aware Reward applied.
% For rollout structure, if the outputs are diversity, we treat this problem as an exploration item and avoid adding Boundary-Aware Reward in this group. Otherwise, the model has a relative high confidence of this problem and IDK should be encouraged. 
% 然而，模型的探索能力和Boundary-Aware的能力总是存在冲突，
% % 对于一个回答问题能力差的模型，其boundary-aware能力会更强。
% 在训练过程中，模型应当随着回答问题能力的提升，不断扩展Reasoning Boundary，然而，直接将Boundary-aware reward引入到模型训练中，会使得模型过多于给出IDK，过早放弃探索，阻碍模型的回答能力的提升。为了实现最佳的exploration-boundary能力的tradeoff，我们从训练阶段和回复结构两个方面去确立balance的准则。
% 为了尽可能避免损害模型的exploration的同时实现模型boundary-awareness，我们从两个方面去做,从训练阶段方面，模型在前期奖励提升倾向于探索，后期奖励提升倾向于，从问题结果方面，一个问题对应的一组rollout回复，如果提取答案的多样性高，说明对于该问题模型的不确定高，还处于探索阶段，反之说明模型对该问题已经处于确定，此时应当鼓励IDK。

% search-o1是否要比较, 
\begin{table*}[t] % [t]表示顶部，*表示跨整栏

\centering
\small % 添加这一行
\begin{tabular}{lcccccccccccc}
\toprule
\multirow{3}{*}{\textbf{Method}} & 
\multicolumn{3}{c}{\textbf{HotpotQA}} & \multicolumn{3}{c}{\textbf{MusiQue}} & \multicolumn{3}{c}{\textbf{2Wiki.}} & \multicolumn{3}{c}{\textbf{Bamboogle}} \\
\cmidrule(lr){2-4} \cmidrule(lr){5-7} \cmidrule(lr){8-10} \cmidrule(lr){11-13}
& Acc & Prec & Rel. & Acc & Prec & Rel. & Acc & Prec & Rel. & Acc & Prec & Rel.  \\
\midrule
% \multicolumn{13}{l}{\textbf{Qwen2.5-7B-Instruct}}\\
% \normalrow
\midrule
\multicolumn{13}{c}{\textbf{\textit{Existing Agentic Search Methods}}} \\\midrule\midrule
Search-R1 & 49.0 & 49.0 & \cellcolor[rgb]{0.92, 0.88, 1.0}49.0 & 22.5 & 22.5 & \cellcolor[rgb]{0.92, 0.88, 1.0}22.5 & 39.0 & 39.0 & \cellcolor[rgb]{0.92, 0.88, 1.0}39.0 & 52.0 & 52.0 & \cellcolor[rgb]{0.92, 0.88, 1.0}52.0
 \\
ReSearch & \textbf{61.5} & 61.5 & \cellcolor[rgb]{0.92, 0.88, 1.0}61.5 & \textbf{31.0} & 31.0 & \cellcolor[rgb]{0.92, 0.88, 1.0}31.0 & 54.2 & 54.2 & \cellcolor[rgb]{0.92, 0.88, 1.0}54.2 & 54.4 & 54.4 & \cellcolor[rgb]{0.92, 0.88, 1.0}54.4 \\
% \hline\hline
\midrule\midrule

% \rowhighlight
\multicolumn{13}{c}{\textbf{\textit{Our Implementations}}} \\
\midrule\midrule
% \hline\hline

Naive RAG & 49.6 & 49.6 & \cellcolor[rgb]{0.92, 0.88, 1.0}49.6 & 12.7 & 12.7 & \cellcolor[rgb]{0.92, 0.88, 1.0}12.7 & 29.5 & 29.5 & \cellcolor[rgb]{0.92, 0.88, 1.0}29.5 & 32.0 & 32.0 & \cellcolor[rgb]{0.92, 0.88, 1.0}32.0
 \\
IRCoT & 52.1 & 52.1 & \cellcolor[rgb]{0.92, 0.88, 1.0}52.1 & 14.2 & 14.2 & \cellcolor[rgb]{0.92, 0.88, 1.0}14.2 & 30.6 & 30.6 & \cellcolor[rgb]{0.92, 0.88, 1.0}30.6 & 36.8 & 36.8 & \cellcolor[rgb]{0.92, 0.88, 1.0}36.8
\\

% Tool-Star* & 61.5 & 61.5 & \cellcolor[rgb]{0.92, 0.88, 1.0}61.5 & 26.0 & 26.0 & 26.0 & \textbf{60.5} & 60.5 & 60.5 & \textbf{59.2} & 59.2 & 59.2 \\

% \midrule

% \multicolumn{13}{l}{\textbf{\textit{Our Implementations}}} \\
TIR Prompt & 51.5 & 51.5 & \cellcolor[rgb]{0.92, 0.88, 1.0}51.5 & 21.5 & 21.5 & \cellcolor[rgb]{0.92, 0.88, 1.0}21.5 & 43.0 & 43.0 & \cellcolor[rgb]{0.92, 0.88, 1.0}43.0 & 48.8 & 48.8 & \cellcolor[rgb]{0.92, 0.88, 1.0}48.8 \\
Reliable TIR Prompt & 52.5 & 62.1 & \cellcolor[rgb]{0.92, 0.88, 1.0}60.6 & 21.0 & 30.0 & \cellcolor[rgb]{0.92, 0.88, 1.0}27.2 & 35.5 & 45.5 & \cellcolor[rgb]{0.92, 0.88, 1.0}43.3 & 47.2 & 50.8 & \cellcolor[rgb]{0.92, 0.88, 1.0}50.5 \\
% Expression Monitor & 29.5 & 60.8 & \cellcolor[rgb]{0.92, 0.88, 1.0}49.2 & 8.5 & 21.3 & 13.6 & 30.5 & 45.1 & 37.7 & 24.0 & 46.9 & 35.4 \\
% Hidden Monitor & \\

\midrule

Reliable RFT & 24.5 & \textbf{68.0} & \cellcolor[rgb]{0.92, 0.88, 1.0}40.2 & 11.0 & 36.6 & \cellcolor[rgb]{0.92, 0.88, 1.0}18.5 & 14.5 & 48.2 & \cellcolor[rgb]{0.92, 0.88, 1.0}23.9 & 36.8 & 56.0 & \cellcolor[rgb]{0.92, 0.88, 1.0}49.4 \\
GRPO & 60.0 & 60.0 & \cellcolor[rgb]{0.92, 0.88, 1.0}60.0 & 29.5 & 29.5 & \cellcolor[rgb]{0.92, 0.88, 1.0}29.5 & 59.5 & 59.5 & \cellcolor[rgb]{0.92, 0.88, 1.0}59.5 & \textbf{57.6} & 57.6 & \cellcolor[rgb]{0.92, 0.88, 1.0}57.6 \\\midrule
\cellcolor[gray]{0.9}BAPO (Ours) & \cellcolor[gray]{0.9}58.0 & \cellcolor[gray]{0.9}66.6 & \cellcolor[rgb]{0.92, 0.88, 1.0}\textbf{65.5} & \cellcolor[gray]{0.9}29.5 & \cellcolor[gray]{0.9}
\textbf{38.8} & \cellcolor[rgb]{0.92, 0.88, 1.0}\textbf{36.6} & \cellcolor[gray]{0.9} 57.0 & \cellcolor[gray]{0.9} \textbf{64.1} & \cellcolor[rgb]{0.92, 0.88, 1.0}\textbf{63.3} & \cellcolor[gray]{0.9}\textbf{57.6} & \cellcolor[gray]{0.9}\textbf{61.5} & \cellcolor[rgb]{0.92, 0.88, 1.0}\textbf{61.2} \\

\bottomrule
\end{tabular}
\caption{Performance on QA tasks with Qwen2.5-7B-Instruct as the backbones. Rel. indicates the reliability metric.}
\vspace{-5pt}

\label{tab:performance}

\end{table*}

\section{Experiments}
\label{sec:experiments}

\subsection{Experimental Setups}
\label{sec:exp-setup}
% To evaluate the effectiveness of ReSearch, we conduct extensive experiments mainly on multi-hop question answering benchmarks that need multi-step reasoning and multiple information retrieval. Our ReSearch is trained from Qwen2.5-7B, Qwen2.5-7B-Instruct, Qwen2.5-32B and Qwen2.5-32B-Instruct [39]. During training, we only use the data from training set of MuSiQue [32], since it has various types of multi-hop questions and constructed via fine-grained quality control.
% \paragraph{Benchmarks.} 
% We use four benchmarks of multi-hop QA tasks: HotpotQA~\cite{hotpotqa}, MuSiQue~\cite{musique}, 2WikiMultiHopQA~\cite{2wiki}, and Bamboogle~\cite{bamboogle}. Specifically, HotpotQA, 2WikiMultiHopQA and MuSiQue are constructed from Wikipedia or Wikidata~\cite{wikidata} using diverse multi-hop mining strategies, while Bamboogle is a manually curated dataset of 2-hop challenging questions.
% Following~\cite{toolstar_dong_2025}, we sample 200 examples of the test set of HotpotQA, 2WikiMultiHopQA, and MuSiQue respectively, and include all 125 test examples from Bamboogle for efficient evaluation.

\paragraph{Baselines.}
We compare BAPO against (1) existing RL agentic search methods, and (2) our implementations of prompting and training-based methods. 
% For existing methods, the details of implementation are listed in Appendix 
For the former, we select two representative approaches: Search-R1~\cite{search-r1} and ReSearch~\cite{research}. 
We use Qwen2.5-7B-Instruct as the backbone for a fair comparison.
The prompt-based baselines include:
(1) Naive RAG: A naive retrieval-based approach that directly concatenates retrieved context with the query.
(2) IRCoT~\cite{ircot}: An interleaving method where retrieval and CoT mutually guide each other.
(3) TIR Prompt~\cite{toolstar_dong_2025}: a standard tool-integrated prompt, which instructs the model to use search tools during reasoning (Appendix~\ref{sec:prompt-template}).
(4) Reliable TIR Prompt: Building on TIR Prompt, it additionally encourages the model to acknowledge ignorance by responding with \textit{IDK} (Section~\ref{sec:impairs}).
For training-based baselines, we choose (1) Reliable Rejection Sampling Fine-Tuning (Reliable RFT), which generates multiple reasoning paths for each question under Reliable TIR Prompt, and incorporates both correct paths and reasonable \textit{IDK} responses into the SFT dataset. 
Similar to BAPO, an \textit{IDK} response is adopted when no correct path is available. 
(2) GRPO~\cite{deepseekmath_shao_2024}, which uses only the correctness reward, with other settings matching those of BAPO.

\begin{table}[t] % [t]表示顶部，*表示跨整栏

\centering
\small % 添加这一行
\setlength{\tabcolsep}{9pt}
\begin{tabular}{lccc}
\toprule
\textbf{Method} & Acc & Prec & Rel. \\
\midrule
\multicolumn{4}{l}{\textbf{Qwen2.5-3B-Instruct}} \\

TIR Prompt & 29.2 & 29.2 & 29.2 \\
Reliable TIR Prompt & 25.4 & 31.8 & 30.2 \\
% Reliable RS & \\
GRPO & \textbf{45.1} & 45.1 & 45.1 \\
\cellcolor[gray]{0.9}BAPO (Ours) & \cellcolor[gray]{0.9}44.8 & \cellcolor[gray]{0.9}\textbf{52.9} & \cellcolor[gray]{0.9}\textbf{51.3} \\

\midrule

\multicolumn{4}{l}{\textbf{Qwen2.5-14B-Instruct}}\\
TIR Prompt & 49.7 & 49.7 & 49.7 \\
Reliable TIR Prompt & 47.6 & 56.7 & 55.0 \\
% Reliable RS & \\

GRPO & \textbf{56.6} & 56.6 & 56.6 \\

\cellcolor[gray]{0.9}BAPO (Ours) & \cellcolor[gray]{0.9}54.0 & \cellcolor[gray]{0.9}\textbf{65.7} & \cellcolor[gray]{0.9}\textbf{63.3} \\

\bottomrule
\end{tabular}
\caption{Performance of BAPO and baselines across different model scales. Rel. indicates the reliability metric.
% The consistent improvements across various benchmarks and model sizes demonstrate the scalability and robustness of our approach.
}\vspace{-10pt}
\label{tab:scale}

\end{table}

\paragraph{Reliability Metric.} 
% To evaluate the correctness of answers, we use LLM-as-a-judge Equal~\cite{llmjudge1} for automatic evaluation as it has a more flexible evaluation criterion to avoid misclassifying semantically equivalent but lexically mismatched answers as incorrect. 
% Following previous work~\cite{research, toolstar_dong_2025}, we extract answers from the model output enclosed in $\textbackslash \text{box}\{\}$, and then use GPT-4 with a judge prompt to measure the correctness of the final answer. 
% More evaluation details and the character-level F1 score and Exact Match (EM) results are shown in Appendix.
% Except for the accuracy and precision metrics defined in Section~\ref{sec:settings}, 
We follow ~\citet{rejectionIR} to adopt a reliability metric to comprehensively balance accuracy and precision:
\begin{equation}
    % \text{reliability} = \lambda \cdot \text{acc} + (1 - \lambda) \cdot \text{prec},
    \text{reliability} = (1- \rho_{\textit{IDK}}) \cdot \text{prec} + \rho_{\textit{IDK}} \cdot \text{acc},
\tag{7}
\end{equation}
% \noindent where $\rho_{\textit{IDK}}$ is the \textit{IDK} rate. This metric represents a natural trade-off between accuracy($\frac{N_c}{N}$) and precision~($\frac{N_c}{N-N_r}$): when $\rho_{\textit{IDK}}$ is low, the model should be more cautious to avoid wrong answers, and when $\rho_{\textit{IDK}}$ is high, the model should be encouraged not to reflexively admit ignorance but rather to attempt to solve the problem. In this way, reliability provides a measurement that aligns with practical user expectations.
where $\rho_{\textit{IDK}}$ denotes the \textit{IDK} rate. 
This metric functions as a dynamic interpolation between precision and accuracy based on the model's tendency to refuse. 
Specifically, when the agent rarely refuses to give an answer (low $\rho_{\textit{IDK}}$), the metric prioritizes precision, enforcing strict correctness on generated answers. 
Conversely, as the agent tends to refuse more frequently (high $\rho_{\textit{IDK}}$), the metric shifts focus toward standard accuracy . This mechanism effecively penalizes excessive ``lazy'' \textit{IDK}.
% encouraging the model to attempt solvable problems rather than reflexively admitting ignorance.
% \noindent where $\lambda \in [0,1]$ indicates user's sensitivity towards misleading answers, while $(1- \lambda)$ measures the demand for helpful answers. We equate $\lambda$ to the answer rate $(1 - \rho_{\textit{IDK}})$ to align with practical user expectations:
% when the model is active to provide certain responses, user become sensitive to errors (requiring high precision); conversely, when model responses are often uncertain, the user's priority shifts towards obtaining any useful information (requiring high accuracy).

\paragraph{Implementation Details.}
% To simplify the training process, we follow the paradigm of directly training the model with RL~\cite{deepseekai2025deepseekr1incentivizingreasoningcapability, r1searcher, li2025torl}
% , eliminating the need for cold-start stage. 
% We utilize a high-quality RL training dataset from \citet{toolstar_dong_2025}, comprising 5000 multi-hop QA samples.

Regarding BAPO-specific hyper-parameters, we set the \textit{IDK} ratio threshold $\alpha$ as 5\%, and the resampling times $k$ is set to 2. 
% A detailed sensitivity analysis of these parameters are presented in Appendix~\ref{sec:sensitivity}.
The retrieval environment is based on FlashRAG~\cite{flashrag2025}. 
We use E5-base-v2~\cite{e5_2024} as the retriever and Wikipedia data~\cite{wikidata} from December 2018 as the knowledge base. 
During training and evaluation, we retrieve top-5 results for each query.
We train the GRPO and BAPO models with a batch size of 64 for 2 epochs.
The rollout size is set to 8, and the maximum tokens are set to 8192. The maximum number of tool invocations during training and inference is set to 3.
Sensitivity analysis of the hyper-parameters are present in Appendix~\ref{sec:sensitivity}.
% More details and sensitivity analysis of the hyper-parameters are present in Appendix~\ref{sec:impl} and Appendix~\ref{sec:sensitivity}.
% (equivalent to 156 training steps). 
% The KL loss coefficient $\beta$ is set to 0.001 and clip ratio $\epsilon$ is set to 0.1. 

\subsection{Main Results}
% 1. 相比于基础的模型
The performance of BAPO and other baselines conducted on Qwen2.5-7B-Instruct is presented in Table~\ref{tab:performance}. Compared to all methods from both existing agentic search and our implemented baselines, 
% For the 7B scale, a comprehensive comparison with all baselines is provided. For the 3B and 14B scales, we focus on three representative baselines that combine ease of implementation with the high reliability demonstrated in our 7B experiments.
% From the results, we can get these key observations:
% \paragraph{BAPO outperforms existing agentic approaches.}
BAPO achieves significant reliability improvements across all benchmarks~(+15.8 scores in average). 
Notably, compared to existing agentic search models such as Search-R1~\cite{search-r1} and ReSearch~\cite{research}, which utilize large-scale training sets of 90k and 19k samples respectively, BAPO-trained model with \textbf{only 5k samples} achieves competitive accuracy and substantially higher reliability. 
% Specifically, BAPO improves 
% consistently outperforms baselines in precision and reliability across four benchmark datasets while maintaining competitive accuracy. 
% When compared to models trained with GRPO, BAPO achieves average improvements in reliability of 13.9\%, 9.7\%, and 11.9\% for the 3B, 7B, and 14B model sizes respectively, with corresponding precision improvements of 17.3\%, 11.8\%, and 16.2\%, while incurring only marginal accuracy decreases of 0.7\%, 2.2\%, and 4.5\%, respectively.

When compared to GRPO, BAPO achieves average improvements in reliability of 9.7\%, with corresponding precision improvements of 11.8\%, while incurring only marginal accuracy decreases of 2.2\%.
These results showcase BAPO's effectiveness and efficiency in training reliable agents.

Regarding methods with reliable techniques, such as Reliable RFT and Reliable TIR Prompt, we find that Reliable RFT tends to be over-conservative; although they achieve significant gains in precision, this come at the cost of a drastic decline in accuracy, suffering a 27-point accuracy drop compared to the TIR Prompt baseline. This ultimately undermines overall reliability. While Reliable TIR Prompt enhances precision without sacrificing accuracy, it inherently lacks the capacity to further improve problem-solving capability. 
On the contrary, BAPO achieves both boundary awareness and problem-solving capability. 
\begin{table}[t]
\small
    \centering
    \setlength{\tabcolsep}{12pt}
    % \small % Uncomment if you need to reduce font size
    \begin{tabular}{cccc}
    \toprule
    Acc & Prec & $\rho_{\textit{IDK}}$ & Reliability \\
    \midrule
    \multicolumn{4}{l}{\cellcolor[gray]{0.93}\textit{BAPO}} \\
    % \midrule
    \textbf{44.8} & 52.8 & 16.8\% & \textbf{51.3} \\
    \midrule
    \multicolumn{4}{l}{\cellcolor[gray]{0.93}\textit{~w/o Boundary-Aware Reward}} \\
    30.6 & \textbf{62.4} & 53.1\% & 44.8 \\  
\midrule
    \multicolumn{4}{l}{\cellcolor[gray]{0.93}\textit{~w/o Sample Modulator}} \\
    43.3 & 52.0 & 20.4\% & 50.1 \\
    \midrule
    \multicolumn{4}{l}{\cellcolor[gray]{0.93}\textit{~w/o Sample \& Stage Modulator}} \\
    37.8 & 56.0 & 35.2\% & 49.0 \\
    \bottomrule
    \end{tabular}
    \caption{Ablation study. The metrics are averaged across four benchmarks on Qwen2.5-3B-Instruct.
    % (HotpotQA, 2Wiki, MuSiQue, Bamboogle). 
    %  
    }\vspace{-10pt}
    \label{tab:ablation_BAPO}
\end{table}

\subsection{Generalization Across Model Scales}
To verify BAPO's generalization across different model scales, we extend our evaluation to the 3B and 14B versions of the Qwen2.5-Instruct series. 
% To verify the generalization of BAPO, we extend our evaluation to the 3B and 14B versions of the Qwen2.5-Instruct series. We include the TIR Prompt and Reliable TIR Prompt to characterize the intrinsic capabilities of the instruct models. Additionally, GRPO is included as a baseline to represent model's potential capability when optimized solely for correctness without boundary constraints.
As shown in Table~\ref{tab:scale}, 
BAPO improves the reliability of its instruct model backbones with Reliable TIR Prompt by an average of 76.1\% and 27.4\% on the 3B and 14B scales, respectively. Furthermore, compared to the standard GRPO baseline, BAPO achieves substantial reliability advantages of 13.9\% and 11.9\%.
These results are consistent with our main experiments, confirming that BAPO is generalizable for building reliable agentic search.

% 更多的cross-scle的道德结论
% Moreover, these results serve as a cross-scale ablation study. The progression from TIR Prompt to + Reliable Prompt, and finally to BAPO, consistently shows that while prompting helps, the RL-based optimization in BAPO is essential for achieving the best trade-off between correctness and boundary awareness. This confirms that BAPO’s effectiveness is not tied to a specific model size but is a generalizable solution for building more reliable agentic search systems.

% For the 3B scale model, which has a relative weak model capability, BAPO significantly improves instruct model's reliability by avergae xxx\%, and compared to model trained with GRPO, BAPO achives an advantage of xxx. 
% For the 14B scale model, BAPO improves instruct model's reliability by average xxx\%, and achieves and advantage of xxx compared to GRPO. 
%上面两段合成一段，通过在强化模型解决问题能力同时保证awareness of reaosning boudnary, BAPO在3b和14b的scale上达到了improves instruct model's reliability by average xxx\% and xxx\%, and achieves and advantage of xxx\% and xxx\% compared to GRPO, respectively. 

% This consistently further amplifies the model's inherent awareness of its , achieving substantial gains in precision while maintaining competitive accuracy.

\subsection{Ablation Study}
% 为了理解每个模块的作用，我们
Table~\ref{tab:ablation_BAPO} details the ablation study on Qwen2.5-3B-Instruct comparing BAPO with its variants. Through the experiments results, we can draw the following observations:
% The full BAPO method achieves the highest accuracy (44.8) and overall reliability (51.3).

\paragraph{Fixed Reward Strategy Leads to Reward Hacking.} 
When replacing our proposed reward mechanism with a fixed positive reward of 0.5 for \textit{IDK} responses, which simulates the setting used in BARREL~\cite{yang2025barrel}, the model exhibits extreme over-conservatism. The \textit{IDK} rate spikes to 53.1\%; while this boosts precision, it drastically reduces accuracy and yields the lowest overall reliability score among all variants.

\paragraph{Adaptive Reward Modulator is Critical.}
Next, we ablate the two components of the Adaptive Reward Modulator to verify their effectiveness. We find that removing sample-level deactivation results in a mild degradation in reliability ($-1.2$). In contrast, removing both the stage-level and sample-level modulators causes the model to degenerate into a local optimum where it exploits \textit{IDK} responses. Consequently, we observe an excessive surge in the \textit{IDK} rate ($16.8\%\rightarrow35.2\%$), accompanied by significant drops in accuracy ($44.8\rightarrow37.8$) and reliability ($51.3\rightarrow49.0$).

\begin{figure}[t] % [t] 表示尽量放在页顶，ACL风格通常推荐这样做
    \centering
    
    % --- 第一个子图 (a) ---
    \begin{subfigure}{\linewidth}
        \centering
        % width设置为0.9\linewidth是为了给左右留点白边，美观一些
        % 这里的 example-image-a 换成你的文件名
        \includegraphics[width=\linewidth]{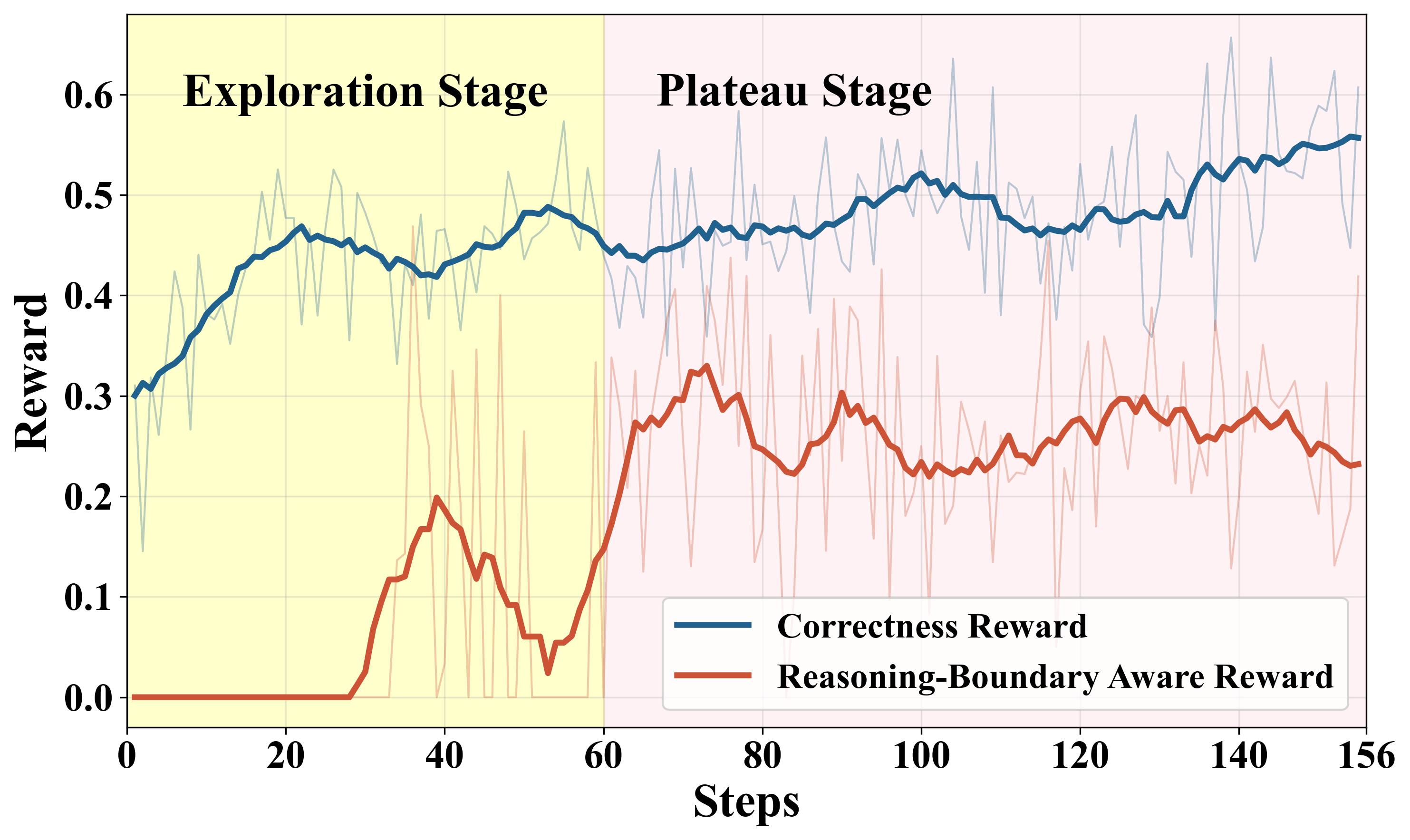}
        \vspace{-8pt}
        % \caption{} % 空的大括号会自动生成 (b)
        \label{fig:sub-first}
    \end{subfigure}
    
    % \vspace{0pt} % 调整两个子图之间的垂直间距
    
    % --- 第二个子图 (b) ---
    \begin{subfigure}{\linewidth}
        \centering
        \includegraphics[width=\linewidth]{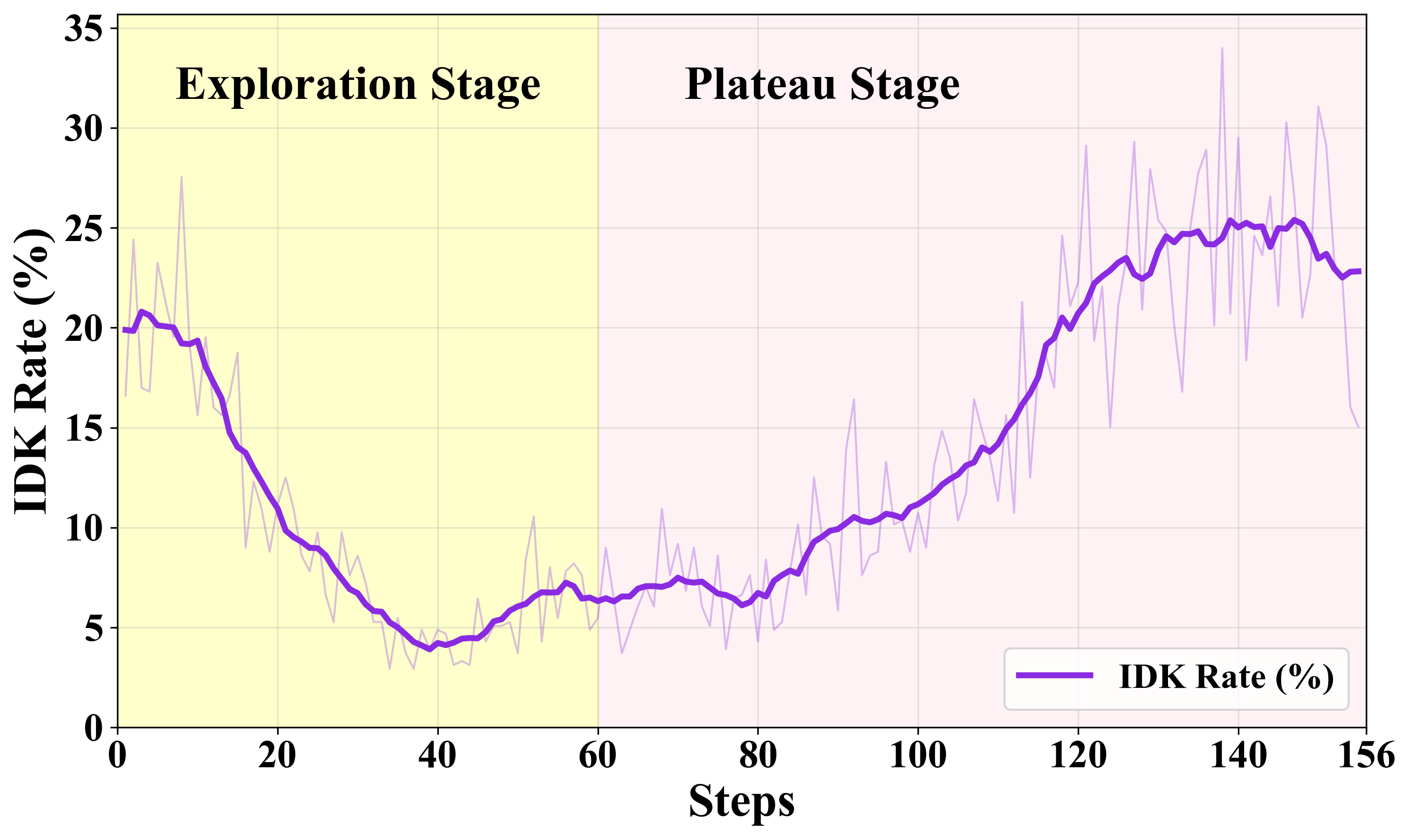}
        \vspace{-15pt}
        % \caption{} % 空的大括号会自动生成 (b)
        \label{fig:sub-second}
    \end{subfigure}
    % \vspace{-15pt}
    % --- 总标题 ---
    \caption{
    % Illustration of 
    \textbf{Upper:} Dynamics of correctness reward $\mathcal{R}^\textit{Correct}$ and boundary-aware reward $\mathcal{R}^\textit{IDK}$. \textbf{Bottom:} Dynamics of \textit{IDK} ratio $\rho_\textit{IDK}$ during BAPO training on Qwen2.5-14B-Instruct.}\vspace{-10pt}
    \label{fig:dynamics}
\end{figure}

\subsection{Reward Dynamics of Two Stage}
%训练过程中的IDK response的占比，已经拒绝问题带来的reward和选择回答的问题的reward对应的动态变化。
% 为了展现训练过程中模型解决问题能力的提升和self-awareness of its reasoning boundary during exploration stage and plateau stage. we visualize the dynamics of basic reward $r_{\mathrm{basic}}$ associated with answered questions and reasoning-boundary aware reward $r_{\mathrm{rba}}$ associated with rejected questions.
% As illustrated in Figure~\ref{fig:dynamics}, during the exploration stage, the basic reward quickly increased and the IDK ratio 也同步下降，这一阶段的模型主要训练自身去solve problem的能力, 当进入到plateau stage, basic reward缓慢提升，同时$r_{\mathrm{rba}}$迅速增加并稳定，带来了IDK ratio的提升。直到后期两歌Reward都逐渐稳定，IDK ratio增长趋缓甚至出现下降。

In Figure~\ref{fig:dynamics}, we visualize BAPO's training dynamics of the average correctness reward $R^{\textit{Correct}}$ and the average boundary-aware reward $R^{\textit{IDK}}$, along with the IDK ratio $\rho_{\textit{IDK}}$ across different stages on Qwen2.5-14B-Instruct.
At the exploration stage (the first 60 steps), $R^{\textit{Correct}}$ increases from 0.3 to 0.5 at the first 60 steps, with $\rho_{\textit{IDK}}$ drops from the initial 20\% to 5\%, indicating that the model prioritizes acquiring problem-solving skills over boundary awareness.
In the plateau stage, the growth of $R^{\textit{Correct}}$ slows down, whereas $R^{\textit{IDK}}$ rises quickly and stabilizes between 0.25 and 0.3. 
This shift drives a sustained increase in the \textit{IDK} rate, eventually restoring $\rho_{\textit{IDK}}$ to over 25\%.

These dynamics revel that our adaptive reward mechanism effectively varies the main optimization goals across training stages, thereby preventing $R^{\textit{Correct}}$ and $R^{\textit{IDK}}$ from hampering each other.
It achieves a superior balance between solving problems and refusing unknown questions. 
% 我们基于stage的动态的奖励机制有效的错开了两个阶段的优化目标，最大程度避免了两种reward的互相矛盾造成的问题，从而
% Finally achieves a good balance between 
% 整体的dynamics的变化趋势，反映了exploration阶段的证实了
% To demonstrate the improvement in the model's problem-solving capabilities and the self-awareness of its reasoning boundary during the exploration and plateau stages, we visualize the dynamics of the correctness reward $R^{\textit{Correct}}$ (associated with answered questions) and the reasoning-boundary aware reward $R^{\textit{IDK}}$ (associated with rejected questions). 
% As illustrated in Figure~\ref{fig:dynamics}, during the exploration stage, the correctness reward increases rapidly while the IDK ratio declines synchronously, indicating that the model primarily focused on acquiring problem-solving skills during this phase. Upon entering the plateau stage, the growth of the correctness reward slows down, whereas $R^{\textit{IDK}}$ rises quickly and stabilizes, driving an increase in the IDK ratio. In the later phase, we notice that the correctness reward and IDK Rate increase simultaneously, indicating a good balance between solving problems and refusing unknown questions. 
% as both rewards gradually stabilized, the growth of the IDK ratio leveled off or even exhibited a slight decline.
% 看一下是否
% (avoiding both excessive rejection and failure to reject).
% GRPO训后的模型答对的问题中，有多少是
% \input{tables/table1}

\begin{figure}[t]
  \centering
  \includegraphics[width=1.0\linewidth]{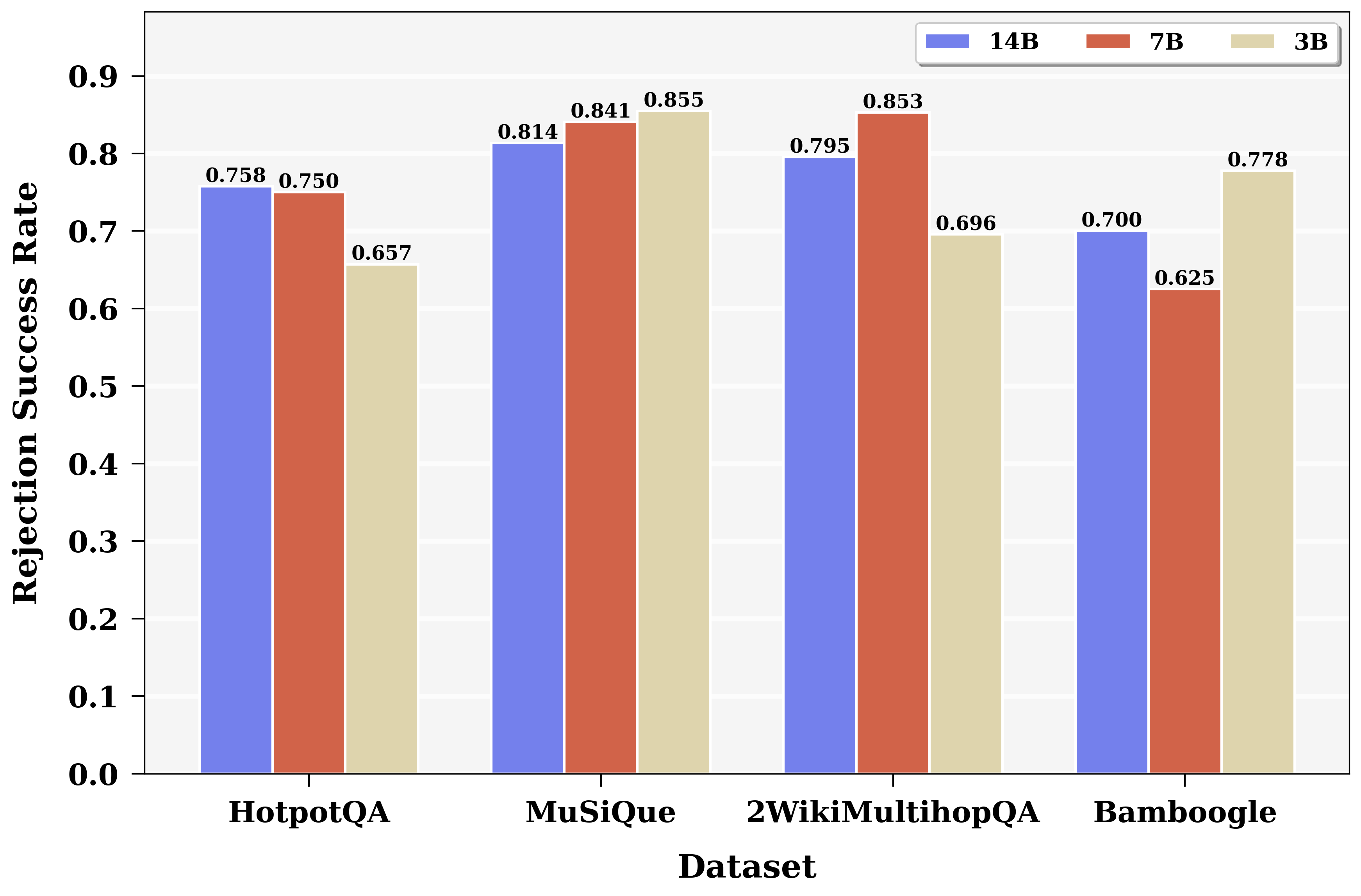} % 图片名/路径
  \caption{Rejection success rates calculated on Qwen2.5-Instruct series models.}\vspace{-10pt}
  \label{fig:rejection_success}
\end{figure}

% 小标题改一下

\subsection{Awareness of When to Refuse}
To assess the rationality of \textit{IDK} responses produced by BAPO, we use \textbf{rejection success rate}, defined as the error rate of the GRPO-trained model on the subset of problems where the BAPO-trained model refuses to give a final answer by responding with \textit{IDK}: 
\begin{equation}
\text{rejection success rate} = \frac{|S^\text{BAPO}_{\text{refuse}} \cap S_{\text{fail}}^{\text{GRPO}}|}{|S^\text{BAPO}_{\text{refuse}}|},
\tag{8}
\end{equation}
where $S^\text{BAPO}_{\text{refuse}} = \{ x \in \mathcal{D} \mid \mathcal{M}_{\text{BAPO}}(x) = \textit{IDK} \}$ and $S_{\text{fail}}^{\text{GRPO}} = \{ x \in \mathcal{D} \mid \mathcal{M}_{\text{GRPO}}(x) \neq \hat{y} \}
$.
To mitigate randomness, we adopt the results of best-of-N (with N=5) as the criterion for determining failures and refusals.

Given that the GRPO-trained model acts as the upper bound of problem-solving capability under our settings, if a problem triggering a refusal is also unsolvable by GRPO, this suggests that the BAPO-trained model's refusal reflects a precise awareness of its reasoning boundaries.
% Concurrently, we present the rejection success rate of random rejection as baseline, which is equivalent to the error rate of GRPO-trained model.
Figure~\ref{fig:rejection_success} shows that the BAPO-trained models attain robust average rejection success rates of 74.7\% (3B), 76.7\% (7B), and 76.7\% (14B). This validates BAPO's ability to enhance model self-awareness, ensuring the appropriate rejection of queries lying outside their reasoning scope. Besides that, it is worth noting that the rejection success rate does not grow with model size, primarily since the overall error rate of the GRPO baselines simultaneously decreases as the model size scales.

% \paragraph{Constructed Unsolvable Question}
% Following~\cite{reliablemath}, we will construct a set of unsolvable adversarial examples and evaluate the performance under these samples.

% \paragraph{Comparison with SFT.}
% Based on rejection sampling, we can construct a set of rejection samples.

% \subsection{Case Study}

% Reliable Prompt可以稳定的带来精准性和幻觉率的改进，但通过其reaosning boundary awareness较低相对于RAS的意识程度来说较低，as its gap between accuracy and precision is limited. 也印证了在合适的奖励设计下，基于强化学习的方法可以进一步加强模型的内在的可靠性对齐。

\section{Conclusion}
In this paper, we reveal a critical challenge that RL training with only correctness rewards 
% can impair the agentic search model's awareness of its reasoning boundaries, causing it to give wrong answers without adequate supporting evidence in the reasoning process rather than admitting \textit{IDK}. However, equipping RL training with rewards for IDK responses leads the model to admit ignorance too readily rather than exploring better solutions during RL, which also 
undermines the model's reliability in providing truthful answers to users. 
To this end, we propose BAPO, a novel RL algorithm for training reliable agentic search models. Beyond correctness rewards, BAPO incorporates boundary-aware rewards to encourage appropriate \textit{IDK} responses. To tackle the tradeoff between exploration and exploitation, we introduce an adaptive reward modulator to prevent the model from being over-encouraged to admit ignorance. 
Extensive experiments show that BAPO achieves superior reliability, enabling the model to maintain awareness of its reasoning boundaries without compromising its capability for deep exploration. 
% to solve hard problems. 
% Furthermore, we provide critical insights for deploying search agents in real-world scenarios and building more trustworthy AI systems.

% \clearpage
\section*{Limitations}
Despite the promising results of BAPO in enhancing agentic search model's reliability, our work has several limitations. 
First, our evaluation primarily focuses on knowledge-intensive tasks. Consequently, the study leaves the generalizability of our method to other types of reasoning problems to be fully explored.
% Furthermore,  due to the high costs associated with commercial search APIs, we only consider a local RAG setup. While this ensures reproducibility, it does not fully replicate the noise, latency, and dynamic nature of real-time web search, which may present additional challenges for reliability.
Furthermore, constrained by computational resources, our experiments currently scale only up to models with 14B parameters. It remains to be seen how the proposed method performs on larger-scale LLMs. Finally, due to the high costs of commercial search APIs, we only consider a local RAG setup. While this ensures reproducibility, it does not fully replicate the noise, and dynamic nature of web search, which may present additional challenges for reliability.

\section*{Acknowledgements}
The project was supported by National Key
R\&D Program of China (No. 2022ZD0160501),
Natural Science Foundation of Fujian Province
of China (No. 2024J011001), and the Public
Technology Service Platform Project of Xiamen
(No.3502Z20231043). We also thank the reviewers
for their insightful comments.


\bibliography{custom}

% \clearpage
\appendix

% these methods are limited to the model's parametric knowledge boundaries, overlooking the tool-use scenarios.
% Although some studies enhance LLM reliability through boundary detection or reliable training , they often struggle to effectively balance accuracy and precision~\cite{rejectionIR, yang2025barrel}.

% In contrast, BAPO is the first exploration of enhancing model boundary awareness within agentic search scenarios via a pure RL approach. By leveraging an adaptive reward mechanism, BAPO effectively optimizes overall reliability.

% \section{Preliminary Study Extension}
% Is this degradation of \textit{IDK} rate comes from the increase in accuracy? 

\clearpage

\section{Implementation Details}
\label{sec:impl}
\subsection{Details of GRPO}
We implement GRPO algorithm based on the verl\footnote{https://github.com/volcengine/verl} framework. The details of training parameters are listed in Table~\ref{tab:hypers}.

\begin{table}[htbp]
  \centering
  \renewcommand{\arraystretch}{1.0}   % 行高微调，可按需删改
  \begin{tabular}{l|c}
    \toprule
    \textbf{Hyper-Parameter} & \textbf{Value} \\
    \midrule
    Learning Rate & 1e-6 \\
    LR Scheduler & Constent \\
    Batch Size & 64 \\
    Mini Batch Size & 16 \\
    Training Epochs & 2 \\
    Number of Rollout & 8 \\
    Rollout Temperature & 1.0 \\
    KL Loss Coefficient & 0.001 \\
    Clip Ratio ($\epsilon$) & 0.1 \\
    Max Tokens & 8192 \\
    \bottomrule
  \end{tabular}
\caption{Training hyper-parameters setting of GRPO.}
\label{tab:hypers}
\end{table}
Consistent with previous agentic search models like Search-R1 and Tool-Star, we exclud all tool invocation results from the loss computation to avoid biasing the policy toward tool outputs. Only tokens from the text-based reasoning process and tool requests are considered during loss calculation.
For fair comparison, all non-BAPO-specific parameters are set to match the standard GRPO configuration.

\subsection{Details of Reliable RFT}
As a variant of Rejection Sampling Fine-tuning (RFT), Reliable RFT aims to move beyond pure accuracy maximization and optimize the model to admit uncertainty in the fine-tuning process.
We use the same training dataset as in the implementation of BAPO. For RFT data construction, we perform rollouts with 8 samples per question using a sampling temperature of 1.0. 
For each question, if no correct answer exists across the rollouts, we adopt the model's \textit{IDK} response (if available); otherwise, we select the correct response(s) for training.
We use LLaMA-Factory\footnote{https://github.com/hiyouga/LLaMA-Factory} training framework for conducting experiments.
The training hyper-parameters are listed in Table~\ref{tab:hypers_rft}.

\begin{table}[htbp]
  \centering
  \renewcommand{\arraystretch}{1.0}
  \begin{tabular}{l|c}
    \toprule
    \textbf{Hyper-Parameter} & \textbf{Value} \\
    \midrule
    Learning Rate & 7e-6 \\
    LR Scheduler & Cosine \\
    Training Epochs & 3 \\
    Warmup Ratio & 0.1 \\
    Batch Size & 8 \\
    Gradient Accumulation Steps & 2 \\
    \bottomrule
  \end{tabular}
  \caption{Training hyper-parameters of Reliable RFT.}
  \label{tab:hypers_rft}
\end{table}

% 1. collecting SFT data, utilize the same training dataset with BAPO's RL stage.
% 2. rollout times=8, sampling_temeprature:1.0 similar to the BAPO's reward setting, for questions that can't sample a correct answer, we adopt its \textit{IDK} responses, otherwise we adopt its correct responses.
% 3. training params.
% learning_rate: 7.0e-6
% num_train_epochs: 3.0
% lr_scheduler_type: cosine
% warmup_ratio: 0.1
% per_device_train_batch_size: 1
% gradient_accumulation_steps: 2

% \subsection{Details of Expression Monitor}
% To implement the Expression Monitor~\cite{ontheselfawareness_2025} for revealing the model's reasoning boundaries, we utilize the set of confident and uncertain expressions shown in Figure~\ref{fig:expmonitor}. These expressions were manually summarized and subsequently verified by an LLM. When the model exhibits more uncertainty than confidence, we classify the corresponding response as \textit{IDK}; otherwise, the original answer is retained for evaluation.

% It is important to note that as a post-processing method, the Expression Monitor does not cultivate model's inherent capability to explicitly admit \textit{IDK}, which often leads to a significant discrepancy between the reasoning process and the final answer.

% \begin{figure}[h]
%   \centering
%   \includegraphics[width=1.0\linewidth]{expmonitor.png} % 图片名/路径
%   \caption{Confident and uncertain expressions.}
%   \label{fig:expmonitor}
% \end{figure}

\subsection{Details of Open-source Models}
For a fair comparison, we directly use the released checkpoint of open-source models trained from Qwen2.5-7B-Instruct and follow the corresponding prompts in the original work to ensure the reproducibility of the results.

%1.训练数据集，训练阶段对应算法
%2.reasoning的prompt有什么特殊的
%3.reward有什么特殊的

\paragraph{Search-R1~\cite{search-r1}}
is a reinforcement learning framework that trains a model to autonomously invoke search engines during the reasoning process. Its reward is determined solely by the correctness of the final answer. The training dataset is a unified dataset merging Natural Questions (NQ) and HotpotQA, amounting to approximately 90k training samples.
\paragraph{ReSearch~\cite{research}}
% is another training framework for enables LLMs to agentic search, the reward function 同时考虑f1 score和format正确性。the training set (19938 samples) of MuSiQue for training.
is another framework designed to enable agentic search capabilities of LLMs. Its reward function incorporates both the prediction F1 score and format correctness. Models are trained using the MuSiQue training set, which comprises 19,938 samples.

% \paragraph{Tool-Star~\cite{toolstar_dong_2025}} 
% integrates six types of tools (including code interpreter and search engine) through a two-stage process: cold-start fine-tuning for initial exploration, and a multi-tool self-critic RL algorithm with hierarchical rewards to optimize tool collaboration. The model is trained on a curated set of 90k mathematical and multi-hop QA samples, with easier samples used for fine-tuning and harder ones reserved for the RL stage.

\subsection{Dataset Details}
For training BAPO, we utilize the high-quality dataset introduced by \cite{toolstar_dong_2025}\footnote{https://huggingface.co/datasets/dongguanting/Multi-Tool-RL-10K}. This training dataset is specifically constructed to be challenging for both Direct Reasoning and Tool-Integrated Reasoning. To target agentic search capabilities, we isolate the QA portion of this dataset. This yields 5,000 training samples derived from HotpotQA and 2WikiMultiHopQA.

\subsection{Benchmark Details}
For evaluation, we use four benchmarks of multi-hop QA tasks: HotpotQA~\cite{hotpotqa}, MuSiQue~\cite{musique}, 2WikiMultiHopQA~\cite{2wiki}, and Bamboogle~\cite{bamboogle}. Following the setup in \citet{toolstar_dong_2025}, the full test set consists of 200 test samples from HotpotQA, 200 from 2WikiMultiHopQA, 200 from MuSiQue, and 125 from Bamboogle.
% the reported results in this section use 80 multi-hop QA pairs, including 20 examples each from each of the test sets, 

% 同样，我们从DRL工具中随机抽取约1万个样本进行强化学习。学习率设置为8，每个样本有8次部署。培训批次总人数为128人，迷你批次人数为16人。最大输出长度设置为4096，以确保探索的完整性。为了加快训练速度，遵循DAPO [77]，我们将训练期间工具调用的最大次数设置为3次。为了稳定训练，我们将GRPO中的基层岭发散系数设为0。强化学习阶段持续两个时代。所有实验均在8块NVIDIA A800显卡上进行。为确保可重复性，所有实验结果均为三次独立试验的平均值。我们在强化学习阶段的验证集中大约有300个样本，包括AIME25、25、MATH500、HotpotQA、2wikiMultiHopQA、MuSiQue和Bamboogle。 值得注意的是，Tool-Star 中所有工具调用结果都是被遮罩的。如图3所示，这些蒙蔽段被排除在损耗计算之外，以防止训练策略偏向工具输出。在损失计算中，仅考虑基于文本的推理过程和工具请求中的令牌。

\section{Additional Experiments}
\subsection{Hyper-Parameters Sensitivity Analysis}
\label{sec:sensitivity}

\begin{table}[t]
    \centering
    % \small % Uncomment if you need to reduce font size
    \begin{tabular}{l|cccc}
    \toprule
    Param. & Acc & Prec & $\rho_{\textit{IDK}}$ & Rel. \\
    \midrule
    % \multicolumn{4}{l}{\cellcolor[gray]{0.85}\textit{BAPO}} \\
    $\alpha$ = 0.0 &40.7 & 40.7 & 0.0\% & 40.7 \\
    
    $\alpha$ = 0.05 & \textbf{44.8} & 52.8 & 16.8\% & \textbf{51.3} \\
    
    % $\alpha$ = 0.1 & 41.3 & 47.5 & 15.1\% & 46.4 \\
    $\alpha$ = 0.2 & 39.2 & \textbf{53.9} & 30.2\% & 49.2 \\
    $\alpha$ = 0.3 & 41.4 & 53.8 & 25.2\% & 50.3 \\
    \midrule
    $k$ = 0 & 43.4 & 50.9 & 15.4\% & 49.0 \\
    
    $k$ = 1 & 44.5 & 50.3 & 13.2\% & 49.2 \\
    
    $k$ = 2 & 44.8 & \textbf{52.8} & 16.8\% & \textbf{51.3} \\
    $k$ = 3 & \textbf{45.1} & 52.7 & 16.3\% & 51.2 \\
    \bottomrule
    \end{tabular}
    % \caption{BAPO xxx. The metrics are averaged across four benchmarks on Qwen2.5-3B-Instruct.
    % % (HotpotQA, 2Wiki, MuSiQue, Bamboogle). 
    % %  
    % }
    \caption{Sensitivity analysis of BAPO's key hyperparameters: \textit{IDK} ratio threshold $\alpha$ and resampling times $k$. Metrics are averaged across four benchmarks (HotpotQA, 2WikiMultihopQA, MuSiQue, Bamboogle) on Qwen2.5-3B-Instruct.}

    \label{tab:paramter_analysis}
\end{table}

To investigate the impact of the values of BAPO's specific hyper-parameters \textit{IDK} ratio $\alpha$ and resampling times $k$, we systematically evaluate model's performance under different hyper-parameter settings, while keeping all other hyper-parameters fixed to the default settings used in the main experiments. All experiments are conducted on Qwen2.5-3B-Instruct, and the results are reported in Table~\ref{tab:paramter_analysis}.

This hyper-parameters analysis reveals these key observations:
(1) The value of $\alpha$ is closely tied to the model’s tendency to acknowledge uncertainty. Since $\alpha$ directly controls the strength of the reward assigned to \textit{IDK} responses, when $\alpha = 0$, the model is never encouraged to produce \textit{IDK} responses in the early stage. As a result, the model has lost the ability to generate \textit{IDK} responses in exploration stage and can no longer learn this behavior in plateau stage. In contrast, larger values of $\alpha$ promote more \textit{IDK} responses during the exploration stage, enabling the model to cultivate boundary awareness.
(2) The resampling times $k$ allows us to dynamically enlarge the effective rollout size by repeatedly sampling within groups that contain no rewardable responses, thereby improving overall reliability. Increasing $k$ from 1 to 2 yields clear gains, while further increasing $k$ to 3 provides only marginal improvement, indicating that moderate resampling times is sufficient.

% 2. resampling times $t$ 可以通过在没有可奖励的组内多次采样，实现动态的rollout size，提高样本训练效率，实现更好的整体的reliability，Further increasing $t$ to 3 provides marginal gains, suggesting that moderate resampling is sufficient.

\begin{figure}[t]
  \centering
  \includegraphics[width=0.9\linewidth]{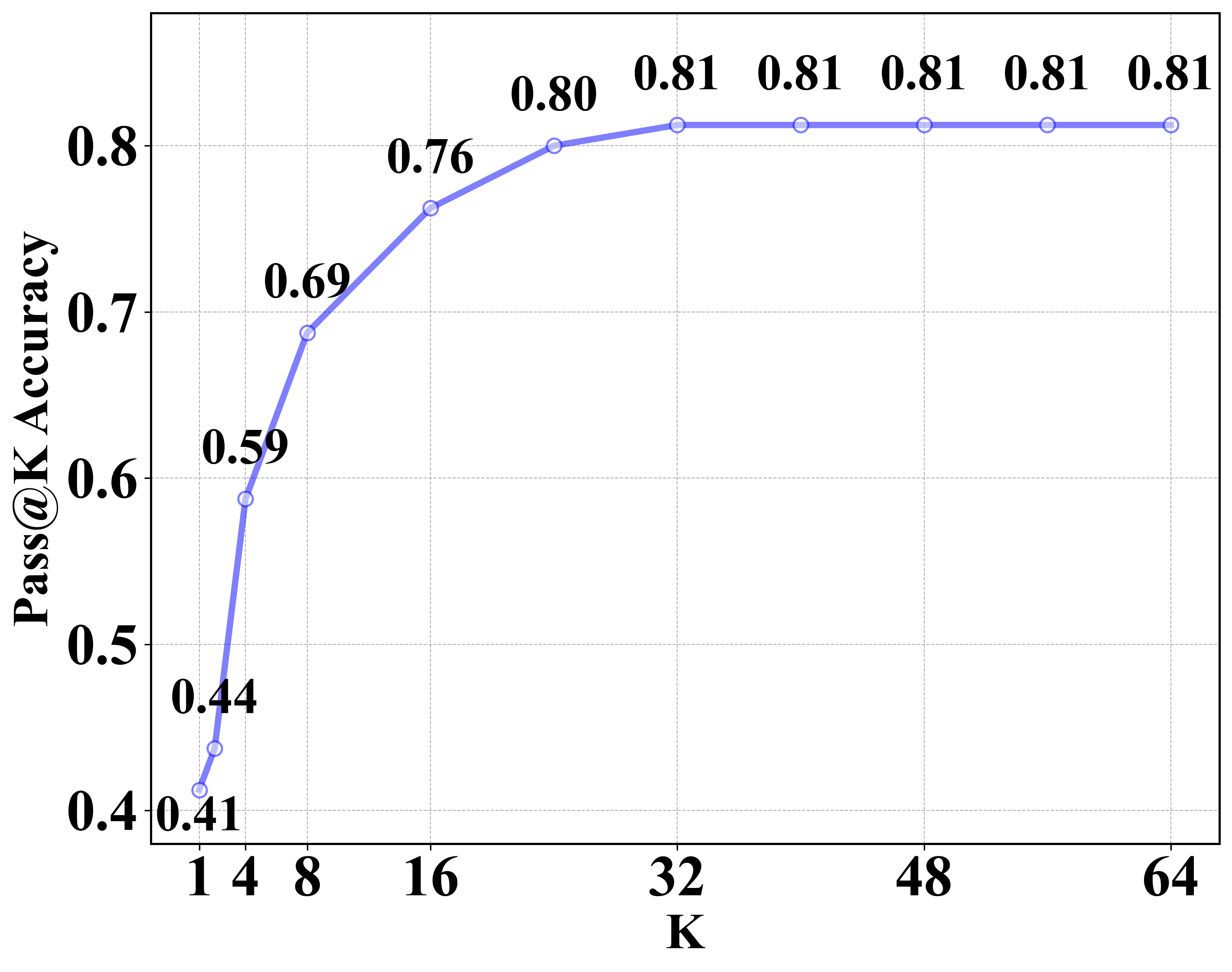} % 图片名/路径
  % \vspace{-5pt}
  \caption{
  Accuracy of the Pass@$K$ on Qwen2.5-7B-Instruct for different values of K. The accuracy improves with increasing $K$ and stabilizes after $K=32$.
}
  \label{fig:passk}
\end{figure}

\subsection{Why We Resample Rollouts}
The resampling strategy is premised on the assumption that increasing the rollout size yields a more accurate estimation of the model's boundary. As illustrated in Figure~\ref{fig:passk}, the model shows no further improvement in pass@$K$ accuracy as $K$ increases to 32. This indicates that failure stems from the problem lying beyond the reasoning boundary, rather than merely being an accidental failure to sample a correct solution. Consequently, our dynamic resampling design adaptively expands the group size to refine this boundary estimation. The results in Table~\ref{tab:paramter_analysis} corroborate our findings, demonstrating that pass@24 to pass@32 (corresponding to resampling factors $k=2$ and $k=3$ with a rollout size of 8) are sufficient for accurate estimation.

\begin{table*}[t] % [t]表示顶部，*表示跨整栏

\centering
\small % 添加这一行
\begin{tabular}{lcccccccccccc}
\toprule
\multirow{3}{*}{\textbf{Method}} & 
\multicolumn{3}{c}{\textbf{HotpotQA}} & \multicolumn{3}{c}{\textbf{MusiQue}} & \multicolumn{3}{c}{\textbf{2Wiki.}} & \multicolumn{3}{c}{\textbf{Bamboogle}} \\
\cmidrule(lr){2-4} \cmidrule(lr){5-7} \cmidrule(lr){8-10} \cmidrule(lr){11-13}
& Acc & Prec & Rel & Acc & Prec & Rel & Acc & Prec & Rel & Acc & Prec & Rel  \\
\midrule
\multicolumn{13}{l}{\textbf{Qwen2.5-7B-Instruct}}\\
% \multicolumn{13}{l}{\textbf{\textit{Open Source Models}}} \\
 %考虑reliable的
Naive RAG & 31.9 & 31.9 & \cellcolor[rgb]{0.92, 0.88, 1.0}31.9 & 6.2 & 6.2 & \cellcolor[rgb]{0.92, 0.88, 1.0}6.2 & 25.8 & 25.8 & \cellcolor[rgb]{0.92, 0.88, 1.0}25.8 & 20.8 & 20.8 & \cellcolor[rgb]{0.92, 0.88, 1.0}20.8 \\
IRCoT & 30.3 & 30.3 & \cellcolor[rgb]{0.92, 0.88, 1.0}30.3 & 6.9 & 6.9 & \cellcolor[rgb]{0.92, 0.88, 1.0}6.9 & 21.6 & 21.6 & \cellcolor[rgb]{0.92, 0.88, 1.0}21.6 & 24.8 & 24.8 & \cellcolor[rgb]{0.92, 0.88, 1.0}24.8
\\
Search-R1 & 39.5 & 39.5 & \cellcolor[rgb]{0.92, 0.88, 1.0}39.5 & 13.5 & 13.5 & \cellcolor[rgb]{0.92, 0.88, 1.0}13.5 & 34.0 & 34.0 & \cellcolor[rgb]{0.92, 0.88, 1.0}34.0 & 40.8 & 40.8 & \cellcolor[rgb]{0.92, 0.88, 1.0}40.8
 \\
ReSearch & \textbf{46.5} & 46.5 & \cellcolor[rgb]{0.92, 0.88, 1.0}46.5 & \textbf{15.5} & 15.5 & \cellcolor[rgb]{0.92, 0.88, 1.0}15.5 &37.5 & 37.5 & \cellcolor[rgb]{0.92, 0.88, 1.0}37.5 & 47.2 & 47.2 & \cellcolor[rgb]{0.92, 0.88, 1.0}47.2 \\
% Tool-Star* & \textbf{46.5} & 46.5 & 46.5 & 10.0 & 10.0 & 10.0 & 38.0 & 38.0 & 38.0 & \textbf{51.2} & \textbf{51.2} & \textbf{51.2} \\

% \midrule

% \multicolumn{13}{l}{\textbf{\textit{Our Implementations}}} \\
TIR Prompt & 19.5 & 19.5 & \cellcolor[rgb]{0.92, 0.88, 1.0}19.5 & 4.0 & 4.0 & \cellcolor[rgb]{0.92, 0.88, 1.0}4.0 & 11.0 & 11.0 & \cellcolor[rgb]{0.92, 0.88, 1.0}11.0 & 26.4 & 26.4 & \cellcolor[rgb]{0.92, 0.88, 1.0}26.4 \\
Reliable TIR Prompt & 24.5 & 29.6 & \cellcolor[rgb]{0.92, 0.88, 1.0}28.3 & 7.0 & 9.5 & \cellcolor[rgb]{0.92, 0.88, 1.0}8.8 & 16.0 & 20.1 & \cellcolor[rgb]{0.92, 0.88, 1.0}19.3 & 25.6 & 27.5 & \cellcolor[rgb]{0.92, 0.88, 1.0}27.4 \\
% Expression Monitor & 15.5 & 24.6 & 21.2 & 3.0 & 5.6 & 4.4 & 8.0 & 11.8 & 10.6 & 17.6 & 27.8 & 24.0 \\
% Hidden Monitor & \\
Reliable RFT & 16.5 & 45.8 & \cellcolor[rgb]{0.92, 0.88, 1.0}27.0 & 5.0 & 16.6 & \cellcolor[rgb]{0.92, 0.88, 1.0}8.5 & 5.5 & 19.6 & \cellcolor[rgb]{0.92, 0.88, 1.0}9.4 & 28.0 & 42.6 & \cellcolor[rgb]{0.92, 0.88, 1.0}37.6 \\
GRPO & 44.0 & 44.0 & \cellcolor[rgb]{0.92, 0.88, 1.0}44.0 & 14.0 & 14.0 & \cellcolor[rgb]{0.92, 0.88, 1.0}14.0 & \textbf{43.5} & 43.5 & \cellcolor[rgb]{0.92, 0.88, 1.0}43.5 & \textbf{48.0 }& 48.0 & \cellcolor[rgb]{0.92, 0.88, 1.0}48.0 \\
\cellcolor[gray]{0.9}BAPO (Ours) & \cellcolor[gray]{0.9}42.0 & \cellcolor[gray]{0.9}\textbf{52.0} & \cellcolor[rgb]{0.92, 0.88, 1.0}\textbf{50.1} & \cellcolor[gray]{0.9}14.0 & \cellcolor[gray]{0.9}
\textbf{20.0} & \cellcolor[rgb]{0.92, 0.88, 1.0}\textbf{18.1} & \cellcolor[gray]{0.9} 42.0 & \cellcolor[gray]{0.9} \textbf{50.2} & \cellcolor[rgb]{0.92, 0.88, 1.0} \textbf{48.6} & \cellcolor[gray]{0.9}46.4 & \cellcolor[gray]{0.9}\textbf{49.5} & \cellcolor[rgb]{0.92, 0.88, 1.0}\textbf{49.3} \\
\midrule
\multicolumn{13}{l}{\textbf{Qwen2.5-3B-Instruct}} \\

TIR Prompt & 7.5 & 7.5 & \cellcolor[rgb]{0.92, 0.88, 1.0}7.5 & 2.0 & 2.0 & \cellcolor[rgb]{0.92, 0.88, 1.0}2.0 & 4.5 & 4.5 & \cellcolor[rgb]{0.92, 0.88, 1.0}4.5 & 11.2 & 11.2 & \cellcolor[rgb]{0.92, 0.88, 1.0}11.2 \\
Reliable TIR Prompt & 16.5 & 22.1 & \cellcolor[rgb]{0.92, 0.88, 1.0}20.7 & 3.5 & 5.2 & \cellcolor[rgb]{0.92, 0.88, 1.0}4.6 & 16.0 & 20.2 & \cellcolor[rgb]{0.92, 0.88, 1.0}19.3 & 16.8 & 17.7 & \cellcolor[rgb]{0.92, 0.88, 1.0}17.6 \\
% Reliable RS & \\
GRPO & \textbf{38.5} & 38.5 & \cellcolor[rgb]{0.92, 0.88, 1.0}38.5 & \textbf{11.0} & 11.0 & \cellcolor[rgb]{0.92, 0.88, 1.0}11.0 & 37.0 & 37.0 & \cellcolor[rgb]{0.92, 0.88, 1.0}37.0 & \textbf{39.2} & 39.2 & \cellcolor[rgb]{0.92, 0.88, 1.0}39.2 \\
\cellcolor[gray]{0.9}BAPO (Ours) & \cellcolor[gray]{0.9}37.0 & \cellcolor[gray]{0.9}\textbf{44.8} & \cellcolor[rgb]{0.92, 0.88, 1.0}\textbf{43.5} & \cellcolor[gray]{0.9}\textbf{11.0} & \cellcolor[gray]{0.9}\textbf{15.9} & \cellcolor[rgb]{0.92, 0.88, 1.0}\textbf{14.4} & \cellcolor[gray]{0.9}36.0 & \cellcolor[gray]{0.9}\textbf{40.0} & \cellcolor[rgb]{0.92, 0.88, 1.0}\textbf{38.9} & \cellcolor[gray]{0.9}\textbf{39.2} & \cellcolor[gray]{0.9}\textbf{42.2} & \cellcolor[rgb]{0.92, 0.88, 1.0}\textbf{42.0} \\

\midrule

\multicolumn{13}{l}{\textbf{Qwen2.5-14B-Instruct}}\\
TIR Prompt & 37.0 & 37.0 & \cellcolor[rgb]{0.92, 0.88, 1.0}37.0 & 9.0 & 9.0 & \cellcolor[rgb]{0.92, 0.88, 1.0}9.0 & 32.5 & 32.5 & \cellcolor[rgb]{0.92, 0.88, 1.0}32.5 & 45.6 & 45.6 & \cellcolor[rgb]{0.92, 0.88, 1.0}45.6 \\
Reliable TIR Prompt & 36.0 & 43.6 & \cellcolor[rgb]{0.92, 0.88, 1.0}42.3 & 12.5 & 16.5 & \cellcolor[rgb]{0.92, 0.88, 1.0}15.5 & 34.0 & 42.5 & \cellcolor[rgb]{0.92, 0.88, 1.0}40.8 & 43.2 & 45.7 & \cellcolor[rgb]{0.92, 0.88, 1.0}45.5 \\
% Reliable RS & \\

GRPO & \textbf{46.0} & 46.0 & \cellcolor[rgb]{0.92, 0.88, 1.0}46.0 & 14.5 & 14.5 & \cellcolor[rgb]{0.92, 0.88, 1.0}14.5 & \textbf{44.5} & 44.5 & \cellcolor[rgb]{0.92, 0.88, 1.0}44.5 & \textbf{53.6} & 53.6 & \cellcolor[rgb]{0.92, 0.88, 1.0}53.6 \\

\cellcolor[gray]{0.9}BAPO(Ours) & \cellcolor[gray]{0.9}45.0 & \cellcolor[gray]{0.9}\textbf{53.8} & \cellcolor[rgb]{0.92, 0.88, 1.0}\textbf{52.3} & \cellcolor[gray]{0.9}\textbf{16.0} & \cellcolor[gray]{0.9}\textbf{22.6} & \cellcolor[rgb]{0.92, 0.88, 1.0}\textbf{20.6} & \cellcolor[gray]{0.9}41.5 & \cellcolor[gray]{0.9}\textbf{53.2} & \cellcolor[rgb]{0.92, 0.88, 1.0}\textbf{50.6} & \cellcolor[gray]{0.9}\textbf{53.6} & \cellcolor[gray]{0.9}\textbf{58.2} & \cellcolor[rgb]{0.92, 0.88, 1.0}\textbf{57.8} \\

\bottomrule
\end{tabular}
\caption{Exact Match (EM) Performance of multi-hop question answering tasks.}
\label{tab:em_performance}

\end{table*}

\subsection{Exact Match Performance}
As a complement of the main experiment  results based on LLM-as-a-Judge Equal (LE), we report the performance under the Exact Match (EM) metric in Table~\ref{tab:em_performance}. Consistent with the LLM-based findings, models trained with BAPO achieve the highest reliability across the 3B, 7B, and 14B model scales. These results further demonstrate the effectiveness and efficiency of BAPO in developing reliable agents.

\begin{table}[t]
    \centering
    % \small % Uncomment if you need to reduce font size
    \begin{tabular}{lcccc}
    \toprule
    Config & Acc & Prec & $\rho_{\textit{IDK}}$ & Rel. \\
    \midrule

    \multicolumn{5}{l}{\cellcolor[gray]{0.93}\textit{Internal Confidence}} \\
    $x=0.2$ & 36.7 & 45.8 & 20.0\% & 43.9 \\
    $x=0.3$ & 34.1  & 48.7 & 30.0\% & 44.3 \\
    $x=0.5$ & 27.5 & 54.8 & 50.0\% & 41.2 \\
    
    \multicolumn{5}{l}{\cellcolor[gray]{0.93}\textit{Expression Certainty}} \\
    $x=0.2$ & 40.4 & 40.4 & 0.0\% & 40.4 \\
    $x=0.3$ & 28.6  & 39.6 & 27.6\% & 36.5 \\
    $x=0.5$ & 26.0 & 43.1 & 38.2\% & 36.6 \\

    \multicolumn{5}{l}{\cellcolor[gray]{0.93}\textit{Self-Reflection}} \\
    % \midrule
    % GPT-4 & \textbf{44.8} & 52.8 & 16.8\% & \textbf{51.3} \\
    - & 17.5 & 56.2 & 68.8\% & 29.5 \\
%     \multicolumn{4}{l}{\cellcolor[gray]{0.93}\textit{Fixed reward for \textit{IDK}}} \\
% 30.6 & \textbf{62.4} & 53.1\% & 44.8 \\
    \bottomrule
    \end{tabular}
    \caption{Performance of uncertainty based methods on Qwen2.5-Instruct-7B, with different values of $x$ .
    % (HotpotQA, 2Wiki, MuSiQue, Bamboogle). 
    %  
    }
    \label{tab:uncertainty-methods}
\end{table}

\subsection{Case Study}
Table~\ref{tab:case-grpo} and Table~\ref{tab:case-BAPO} represent the reasoning process of Qwen2.5-14B-Instruct trained with GRPO and BAPO, respectively.
When the retrieved results lack relevant information, the GRPO-trained model initially identifies the absence of necessary information and attempts to search for it. However, it subsequently hallucinates that the director is Ralph Pappier, resulting in an erroneous final outcome.
In contrast, BAPO enables the model to recognize its boundary. When no relevant information about the film ``Winds of the Pampas'' is found, the model trained with BAPO appropriately responds with \textit{``I DON'T KNOW''} (\textit{IDK}).

\subsection{Detailed Analysis of Uncertainty Estimation Methods}
Although uncertainty estimation methods have proven effective in standard QA or math reasoning tasks~\cite{quantifyuncertain_chen_2023, kuhn2023semanticuncertaintylinguisticinvariances, ontheselfawareness_2025}, their effectiveness remains unexplored in agentic search scenario.
In this section, we systematically evaluate three distinct uncertainty estimation methods: 1) \textbf{Internal Confidence} based on token probabilities, 2) \textbf{Expression Certainty} measuring confidence via the model's verbalized expressions, and 3) \textbf{Self-Reflection} utilizing the model's self-verification capabilities. The implementation details are as follows:
\begin{itemize}
    \item \textbf{Internal Confidence.} Model confidence is conventionally quantified via the probability distribution of its output~\cite{kuhn2023semanticuncertaintylinguisticinvariances, kumar2024confidencehoodinvestigationconfidenceprobability}. In our approach, we adopt the average log-probability as the proxy for internal confidence.
    \item \textbf{Expression Certainty.} To assess uncertainty based on model's expression, we follow \citet{ontheselfawareness_2025} to calculate the proportion of confident versus uncertain expressions within the reasoning trajectory. Specifically, we utilize a predefined lexicon of confident and uncertain phrases (as illustrated in Figure~\ref{fig:expmonitor}), which are manually curated and subsequently verified by an LLM. We quantify the degree of certainty as the ratio of confident expressions to uncertain expressions. 

    \begin{figure}[h!]
  \centering
  \includegraphics[width=1.0\linewidth]{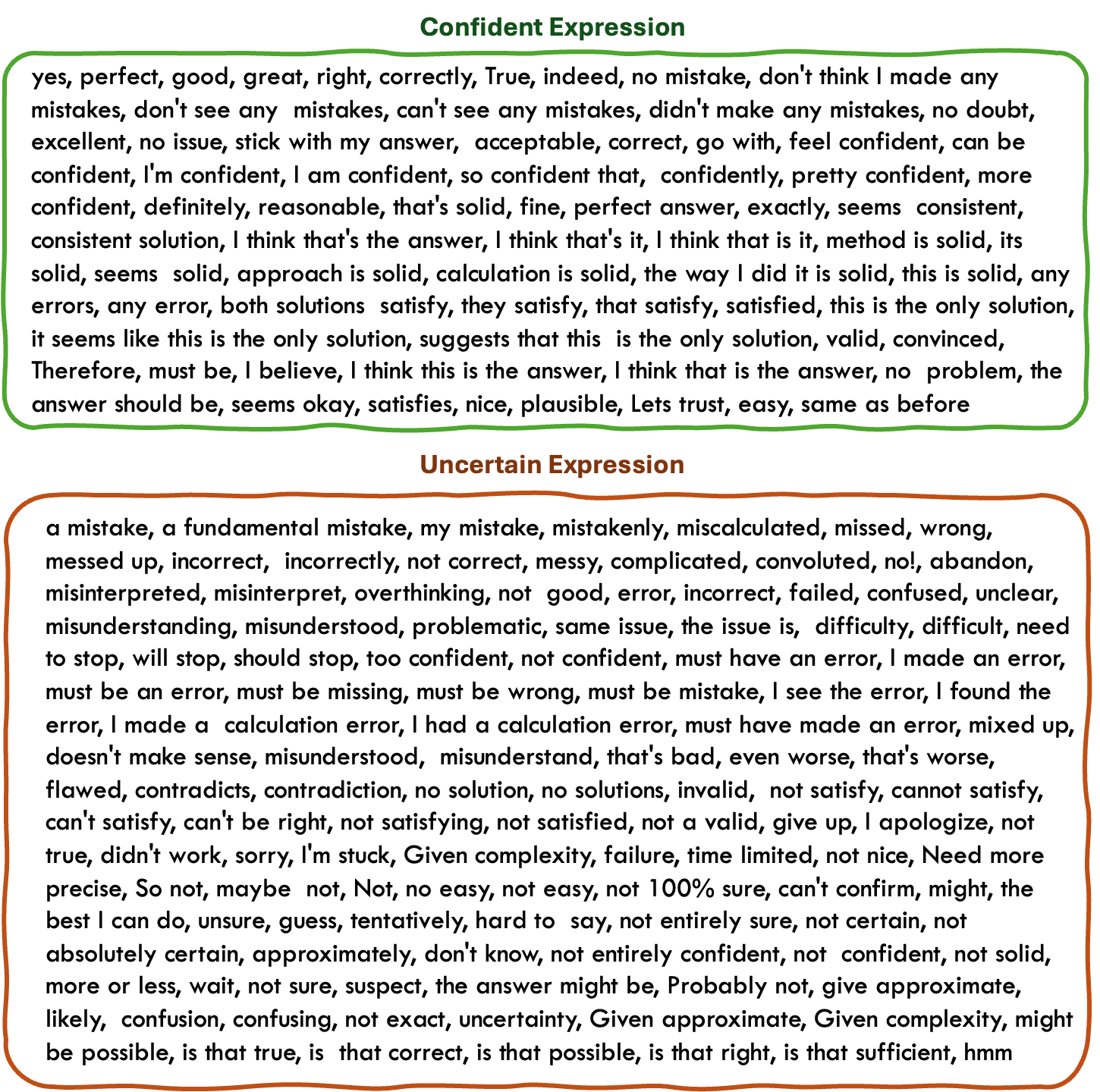} % 图片名/路径
  \caption{Confident and uncertain expressions.}
  \label{fig:expmonitor}
\end{figure}

    \item \textbf{Self-Reflection.} Leveraging the inherent capacity of LLMs for self-reflection~\cite{selfcheckgpt_manakul_2023, madaan2023selfrefineiterativerefinementselffeedback}, we use a prompt-based self-reflection approach to instruct the model to retrospectively audit its own reasoning trajectory to identify potential logical pitfalls. The specific prompt utilized for this verification is detailed in Section~\ref{sec:self-reflection-template}. 
\end{itemize}

% \paragraph{Internal Confidence.}

% \paragraph{Expression Certainty.}

% \paragraph{Self-Reflection.}

For Internal Confidence and Expression Certainty, we apply a percentile-based thresholding method: responses ranked within the bottom $x\in[0,1]$\ of confidence scores across the dataset are classified as \textit{IDK}. For Self-Reflection, we rely on the model's judgment, where any response flagged as uncertain by the LLM is directly labeled as \textit{IDK}. 

As shown in Table~\ref{tab:uncertainty-methods}, although employing these methods improves precision, it comes at the cost of a substantial drop in accuracy, resulting in negligible gains or even degradation in reliability. This underscores a critical challenge in agentic search scenario: the validity of reasoning is inextricably linked to the quality of retrieved content. This external dependency induces a decoupling between the actual reachability of the answer and the model's internal or explicit uncertainty signals, thereby limiting the effectiveness of standard estimation paradigms.
It is important to note that uncertainty estimation methods do not cultivate model's inherent capability to explicitly admit \textit{IDK}, which often leads to a significant discrepancy between the reasoning process and the final answer.

% 对于前两种方法，我们将置信度处于后x\%的回答设置为\textit{IDK}, 对于self-reflection，我们将LLM最后认为uncertain的回答设置为\textit{IDK}. 最后的评估结果在Table~\ref{tab:uncertainty-methods}.
% Responses that are flagged as uncertain or flawed during this check are systematically reclassified as \textit{IDK}.

% \textbf{Internal Confidence:} 模型预测的置信度通常被视作...~\cite{kuhn2023semanticuncertaintylinguisticinvariances, kumar2024confidencehoodinvestigationconfidenceprobability}, 这里我们采用了模型输出内容的log-probability作为置信度指标。

% \paragraph{Expression Certainty:} For measure model's uncertainty based on its expression, we follow \cite{ontheselfawareness_2025} to calculate the confident expression and uncertain expression ratio in reasoning trajectory;
% Specifically, we utilize the set of confident and uncertain expressions shown in Figure~\ref{fig:expmonitor}. These expressions were manually summarized and subsequently verified by an LLM. uncertainty的程度可以视为不确定表达/确定性表达的ratio. 

% 我们通过后处理的方式，将置信度处于后x\%的回答答案设置为IDK。When the model exhibits more uncertainty than confidence, we classify the corresponding response as \textit{IDK}; otherwise, the original answer is retained for evaluation.

% Additionally, 基于LLM具有自我反思，自我检查的能力~\cite{selfcheckgpt_manakul_2023, madaan2023selfrefineiterativerefinementselffeedback}, prompting the models to reflect on their own response to identify potential mistakes is another line of direction to uncertainty estimation. 具体的prompting在Sectionx中展示，我们将check结果为uncertainty的回复置为IDK.

% Additionally, we also evaluate LLM-Checker based method, we 

% 实验结果在Table

%

% 先列一个表

% \clearpage

\begin{table*}[htbp]
  \begin{center}
  \begin{tabular}{p{0.95\textwidth}}
  \toprule
  \textbf{Question:}What is the place of birth of the director of film Winds Of The Pampas?  \\
  \textbf{Answer:} Italian \\
  \midrule 
  \thinkrow{To answer this question, I need to find out who the director of the film "Winds of the Pampas" is and then find their place of birth. I will start by searching for the director of the film "Winds of the Pampas".}
  \searchrow{director of film Winds Of The Pampas}
  \resultrow{
  "The Wind Journeys" The Wind Journeys The Wind Journeys () is a 2009 Colombian-German-Argentine-Dutch drama film written and directed by Ciro Guerra... ...

  "Ralph Pappier" Ralph Pappier Ralph Pappier ( 16 January 1914 in Shanghai – 29 August 1998 in Buenos Aires) was an Argentine production designer, set decorator and film director... ...

  "Savage Pampas (1945 film)" Savage Pampas (1945 film) Savage Pampas (Spanish:Pampa bárbara) is a 1945 Argentine historical film directed by Lucas Demare ... ...
}
  
  \thinkrow{From the search result, I can see that the film "Winds of the Pampas" is not directly mentioned. However, I found that Ralph Pappier is an Argentine film director who has directed several films. I will check if "Winds of the Pampas" is one of his works. If not, I will look for the director of "Winds of the Pampas" again.}
  
  \searchrow{Winds of the Pampas director}
  \resultrow{
  "Ralph Pappier" Ralph Pappier Ralph Pappier ( 16 January 1914 in Shanghai – 29 August 1998 in Buenos Aires) was an Argentine production designer, set decorator and film director ... ...
  
  "Pampa Ilusión" Pampa Ilusión Pampa Ilusión (Illusion of Pampa) is a Chilean telenovela produced by TVN. It was written by Víctor Carrasco, Larissa Contreras, María José Galleguillos and Alexis Moreno. Directed by Vicente Sabatini ... ...
  
  "Law of the Pampas" Law of the Pampas Law of the Pampas is a 1939 American Western film directed by Nate Watt and written by Harrison Jacobs ... ...
  % "Víctor Erice" Víctor Erice Víctor Erice Aras (; born 30 June 1940) is a Spanish film director. He is best known for his films, ""The Spirit of the Beehive"" (1973), that many regard as one of the greatest Spanish films ever made, and ""El Sur"" (1983). Erice was born in Karrantza, Biscay. He studied law, political science, and economics at the University of Madrid. He also attended the Escuela Oficial de Cinematografia in 1963 to study film direction. He wrote film criticism and reviews for the Spanish film journal ""Nuestro Cine"", and made a series of short films before making his first
  
  % "Juan José Jusid" Juan José Jusid Juan José Jusid (born September 28, 1941) is an Argentine film director and screenwriter. Jusid was born in Buenos Aires. He started his professional career as an actor, puppeteer and stage photographer in the 1960s then switched to film studies at the Association of Short Film Directors. He turned director and screenwriter in 1968 and has directed acclaimed films such as ""Bajo Bandera"" (1997), ""Un Argentino en New York"" (1998) and ""Apasionados"" (2002). His films starring actor Miguel Ángel Solá have won a number of Silver Condor awards such as ""Asesinato en el senado de la nación"" ... ...
  }
  \thinkrow{\textcolor{red}{From the search result, I can see that Ralph Pappier is the director of "Winds of the Pampas".} Now, I need to find his place of birth.}
  \searchrow{Ralph Pappier place of birth}
  \resultrow{
  "Ralph Pappier" Ralph Pappier Ralph Pappier ( 16 January 1914 in Shanghai – 29 August 1998 in Buenos Aires) was an Argentine production designer, set decorator and film director... ...
  
  "Ralph Pappier" film ""Circus cavalcade"". The Argentine Academy of Cinematography Arts and Sciences gave him awards for Best Scenography for ""En el viejo Buenos Aires"" (1942) and ""Su mejor alumno"" (1944)... ...
  }
  \thinkrow{From the search result, I can see that Ralph Pappier was born in Shanghai, China.}
  \rowcolor{yellow!10}   
  <answer> The final answer is \verb|\boxed{Shanghai, China}| </answer> \\
  % \bottomrule
  \end{tabular}
  \end{center}
    \caption{A case study on Qwen2.5-14B-Instruct trained with GRPO. Text in \textcolor{red}{red} indicates that the model is hallucinating and providing a conclusion without relevant information in retrieval results.}
\label{tab:case-grpo}

\end{table*}

% Table2——BAPO
\begin{table*}[htbp]
  \begin{center}
  \begin{tabular}{p{0.95\textwidth}}
  \toprule
  \textbf{Question:}What is the place of birth of the director of film Winds Of The Pampas?  \\
  \textbf{Answer:} Italian \\
  \midrule 
  \thinkrow{To answer this question, I need to find out the director of the film "Winds Of The Pampas" and then find out their place of birth. I will start by searching for information on the film "Winds Of The Pampas"}
  \searchrow{Winds Of The Pampas film}
  \resultrow{
  "Savage Pampas (1945 film)"
  Savage Pampas (1945 film) Savage Pampas (Spanish:Pampa bárbara) is a 1945 Argentine historical film directed by Lucas Demare and Hugo Fregonese and starring Francisco Petrone, Luisa Vehil and Domingo Sapelli... ...
  
  "Law of the Pampas" 
  Law of the Pampas Law of the Pampas is a 1939 American Western film directed by Nate Watt and written by Harrison Jacobs. The film stars William Boyd, Russell Hayden, Sidney Toler, Steffi Duna, Sidney Blackmer, Pedro de Cordoba and William Duncan... ...
  
  "Savage Pampas (1966 film)"
  Savage Pampas (1966 film) Savage Pampas is a 1966 western film directed by Hugo Fregonese and starring Robert Taylor, Ron Randell and Marc Lawrence... ...
}
  
  \thinkrow{From the search result, I couldn't find any information about the film "Winds Of The Pampas". However, I found information about other films set in the Pampas region, such as "Savage Pampas" (1945 and 1966) and "Pampa Ilusión". Since there is no information about "Winds Of The Pampas", I will have to conclude that the film may not be well-known or may not have a lot of information available online. Therefore, I don't have enough information to answer the question.}

  \rowcolor{yellow!10}   
  <answer>\verb|\boxed{I DON'T KNOW}| </answer> \\
  % \bottomrule
  \end{tabular}
  \end{center}
    \caption{A case study on Qwen2.5-14B-Instruct trained with BAPO.}
  \label{tab:case-BAPO}

\end{table*}

\section{Prompt Template}
\label{sec:prompt-template}

% \clearpage

\subsection{TIR System Prompt}
\label{sec:tir-prompt}
Table~\ref{tab:tir-prompt} presents the Tool-Integrated Reasoning Prompt for formulating the reasoning process and the search tool call format.

\subsection{Prompt for LLM-as-a-Judge Evaluation}
\label{sec:llm-judge-template}
Table~\ref{tab:llm-prompt} presents the evaluation prompt for LLM-based scoring. It guides the judge model to determine the equivalence of the predicted and labeled answers.

\subsection{Prompt for Self-Reflection}
\label{sec:self-reflection-template}
Table~\ref{tab:reflection-prompt} details the Self-Reflection Prompt, which tasks the model with identifying logical inconsistencies within its reasoning trajectory. 

\section{Detailed Related Work}

\subsection{Agentic Search}
% For time-sensitive or knowledge-intensive questions, Retrieval-Augmented Generation (RAG) enables LLMs to access external search systems rather than relying solely on parametric knowledge, thereby improving response quality~\cite{rag_lewis2020, rag_gao2023, rag_fan2024}.
% While some RAG approaches focus on sophisticated prompting strategies to guide LLMs through question decomposition, query generation, and sub-question answering~\cite{selfrag, searcho1}, these methods typically depend on intricate predefined workflows and have limited effectiveness.
% Another line of research focuses on developing these capabilities through supervised fine-tuning (SFT)~\cite{ragit,corag, sft1_cognitivekernelproframeworkdeep} or DPO~\cite{ragddr}. 
% However, these methods rely on complex data construction pipelines, restricting their generalization ability in new scenarios.

To enhance response quality in knowledge-intensive scenarios, Retrieval-Augmented Generation (RAG)~\cite{rag_lewis2020, rag_gao2023,zhang2025faithfulrag} supplements LLMs' parametric knowledge with external search~\cite{rag_fan2024,zhang2025survey,xiang2025use,zhuang2025linearrag}. Current prompting-based strategies often depend on predefined workflows for interleaving reasoning and search~\cite{selfrag, searcho1}, resulting in limited effectiveness. 
While learning-based methods (SFT/DPO) offer an alternative~\cite{ragit,corag, sft1_cognitivekernelproframeworkdeep, ragddr,li2025mol}, they remain constrained by complex data pipelines and suboptimal generalization.

Inspired by the success of reinforcement learning on mathematics and coding \cite{openaio1,deepseekai2025deepseekr1incentivizingreasoningcapability}, RL-based agentic search systems have emerged, such as Search-R1~\cite{search-r1}, ReSearch~\cite{research}, R1-Searcher~\cite{r1searcher}, and Tool-Star~\cite{toolstar_dong_2025}.
These methods employ format and outcome correctness rewards to enable autonomous query decomposition and flexible interaction with external search systems during reasoning, thereby achieving superior accuracy on complex queries.
Nevertheless, current methods prioritize search accuracy while ignoring holistic reliability, which depends on both accuracy and precision. 
In contrast, our proposed BAPO leverages boundary-aware reward mechanisms to optimize overall reliability, thereby maximizing the practicality of agentic search models.

\subsection{LLM Reliability}
% LLM reliability refers to a model's ability to deliver reliable outputs across interaction scenarios, fundamentally representing the alignment between model behavior and its knowledge boundaries~\cite{enhancing_zheng_2025}. 
% A reliable search system should maximize answer accuracy as much as possible while rejecting out-of-scope queries to avoid generating misleading responses.
One established approach to enhancing LLM reliability involves uncertainty-based methods, which quantify prediction confidence through token probabilities and consistency across diverse outputs \cite{selfcheckgpt_manakul_2023, quantifyuncertain_chen_2023}. 
Alternatively, calibration strategies such as prompting \cite{just_tian_2023} and fine-tuning \cite{whentotrust_tao_2024} are employed to align model confidence with accuracy. 
Recently, \citet{ontheselfawareness_2025} find that reasoning models exhibit self-awareness regarding their capability boundaries and proposed test-time monitoring strategies to terminate unproductive reasoning for unsolvable questions.
Distinctly, our work focuses on activating the model's intrinsic boundary awareness through RL, which is orthogonal to these uncertainty-based methods.

% A parallel line of research focuses on improving a model's internal capacity for refusal.
% 现有的一些工作同样致力于提升模型本身
Sharing this focus on model's internal capacity to refuse queries out of its answerable boundary,
~\citet{rejectionIR}, \citet{teaching_chen_2024} and \citet{sun2025dividethenalignhonestalignmentbased} develop specialized reward models by constructing preference pairs of refusal and standard responses for subsequent RL training.
KnowRL~\cite{knowrl_ren_2025} incorporates a fact-level verification reward into RL to improve factual accuracy, yet it remains dependent on manually curated factual databases.
BARREL~\cite{yang2025barrel} fine-tunes models on distilled reasoning traces that align with expected patterns, followed by RL training with a static medium-level reward for uncertain responses. 
However, our empirical findings suggest that such static reward leads to excessively high rejection rates, thereby impairs accuracy.
Unlike prior work, our approach targets the agentic search scenario, which is characterized by a sophisticated synergy between reasoning and search tool interaction.

\subsection{Policy Optimization}
% DeepSeek-R1~\cite{deepseekai2025deepseekr1incentivizingreasoningcapability} shows that RL with formatting and result-only rewards can steer LLMs
% toward complex chain-of-thought reasoning, boosting performance on challenging tasks.
% Subsequently, Search-R1~\cite{search-r1}, TORL~\cite{li2025torl} explored extending the LLM-R1 approach to tool use.
% However, various RL的变种被提出为了适应多样化的训练要求 to adaptively 调整训练目标和训练难度~\cite{yang2026alignmentmetaweightedonlinesampling, CLPO,tspo,topo}. 
% Different from these previous work, our proposed BAPO aims at balancing two 冲突的reward, 提出了以分阶段为基础的adaptive reward modulator. 通过训练实现模型的boundary-awareness和problem-solving能力的平衡。
DeepSeek-R1~\cite{deepseekai2025deepseekr1incentivizingreasoningcapability} shows that RL with formatting and result-only rewards can steer LLMs toward complex chain-of-thought reasoning, boosting performance on challenging tasks. Subsequently, Search-R1~\cite{search-r1} and TORL~\cite{li2025torl} explored extending this R1-style approach to tool use. However, to accommodate diverse training requirements, various RL variants have been proposed to adaptively adjust training objectives and task difficulty~\cite{CLPO,tian2025seear1treestructuredreinforcementfinetuning,yang2026alignmentmetaweightedonlinesampling,tspo,topo}. Unlike prior work, our proposed BAPO aims to balance two conflicting rewards by introducing a stage-based adaptive reward modulator, thereby achieving an equilibrium between the model's boundary awareness and problem-solving capabilities during training.

% For the aim of balance various reward. 
% to mitigate mode collapse and prevent the policy from converging to a narrow set of interaction patterns, introduce a structural balance reward~\cite{topo}.

\section{The Use of Large Language Models}

In preparing this paper, we made limited use of Large Language Models (LLMs). Specifically, LLMs were employed for two purposes: (i) to aid in polishing the writing by improving grammar, readability, and clarity without altering the scientific content, and (ii) to assist in retrieval and discovery tasks, such as identifying and organizing related work. No LLMs were used for generating novel research ideas, designing experiments, or analyzing results. All conceptual and technical contributions presented in this paper are the sole work of the authors.

% \clearpage
\begin{table*}[htbp]
\begin{tcolorbox}[colback=gray!10, colframe=black!100, title=TIR System Prompt, width=\textwidth]
You are a helpful assistant that can solve the given question step by step with the help of the wikipedia search tool.\\
Given a question, you need to first think about the reasoning process in the mind and then provide the answer.
During thinking, you can invoke the wikipedia search tool to search for fact information about specific topics if needed.\\
The reasoning process and answer are enclosed within \texttt{<think> </think>} and \texttt{<answer> </answer>} tags respectively,
and the search query and result are enclosed within \texttt{<search> </search>} and \texttt{<result> </result>} tags respectively.\\
For example, \texttt{<think>} This is the reasoning process. \texttt{</think>} \texttt{<search>} search query here \texttt{</search>} \texttt{<result>} search result here \texttt{</result>} \texttt{
<think>} This is the reasoning process. \texttt{</think> <answer>} The final answer is \texttt{[ \textbackslash
boxed\{answer here\} ] </answer>}.\\
In the last part of the answer, the final exact answer is enclosed within \texttt{\textbackslash
boxed\{\}}.
\end{tcolorbox}
\caption{Tool-Integrated-Reasoning System Prompt}
\label{tab:tir-prompt}

\end{table*}

\begin{table*}[htbp]

\begin{tcolorbox}[colback=gray!10, colframe=black!100, title=LLM-as-a-Judge Evaluation, width=\textwidth]
You are an evaluation assistant. Please determine if the model output is equivalent to the labeled answer.
\\
Question: \{question\}
\\
Labeled Answer: \{labeled answer\}
\\
Model Output: \{pred answer\}
\\
Did the model give an answer equivalent to the labeled answer? Please respond with "Correct" if they are equivalent, or "Incorrect" if they are not equivalent. Do not include any other text.
\end{tcolorbox}
\caption{LLM-as-a-Judge Evaluation Prompt}
\label{tab:llm-prompt}

\end{table*}

\begin{table*}[htbp]

\begin{tcolorbox}[colback=gray!10, colframe=black!100, title=Self-Reflection Prompt, width=\textwidth]
Below is the reasoning trajectory of an agentic search model. \texttt{<search>}...\texttt{</search>} indicates search queries executed via tools, \texttt{<result>}...\texttt{</result>} shows the returned search results, and \texttt{<answer>}...\texttt{</answer>} contains the model's final answer.

Analyze the following reasoning trajectory for logic errors or inconsistencies.

Reasoning Trajectory:
\{reasoning trajectory\}

Please determine if the response exhibits any of the following issues:\\
1. Logical contradictions or self-contradictions\\
2. Obvious flaws in the reasoning process\\
3. Discrepancy between the answer and the reasoning process
\\
\\
Please answer only "Yes" or "No" without explanation. If any of the above problems exist, answer "Yes"; if there are no obvious issues, answer "No".
\end{tcolorbox}
\caption{Self-Reflection Prompt}
\label{tab:reflection-prompt}

\end{table*}

% \subsection{Prompt for Naive RAG}
% \label{sec:naive-rag}
% \begin{tcolorbox}[colback=gray!10, colframe=black!100, title=LLM-as-a-Judge Evaluation, width=\textwidth]
% Find the useful content from the provided documents, then answer the question.
% Answer the question directly. 
% Your response should be very concise. 
% Please provide use `So the final answer is:’ as a prefix for the final answer. \\ 
% The following are given documents. \\
% Documents: \{reference\} \\
% Question: \{question\}
% \end{tcolorbox}

\end{document}